\documentclass[]{mosi}

\usepackage[table]{xcolor}
\usepackage{booktabs}
\usepackage{multirow}
\usepackage{makecell}
\usepackage{colortbl}
\usepackage{wrapfig}
\usepackage{graphicx}
\usepackage{amsmath}
\usepackage{longtable}
\usepackage{array}
\usepackage{amssymb}
\usepackage[linesnumbered,ruled,vlined]{algorithm2e}
\usepackage{appendix}
\usepackage{makecell}

\definecolor{oursgray}{gray}{0.95}

\definecolor{MossCyan}{HTML}{82D9FF} 
\definecolor{MossBlue}{HTML}{82B1FF}

\definecolor{tickG}{HTML}{00C853}
\definecolor{crossR}{HTML}{FF1744}

\newcommand{\faHome}{\raisebox{-0.2ex}{\includegraphics[height=2.0ex]{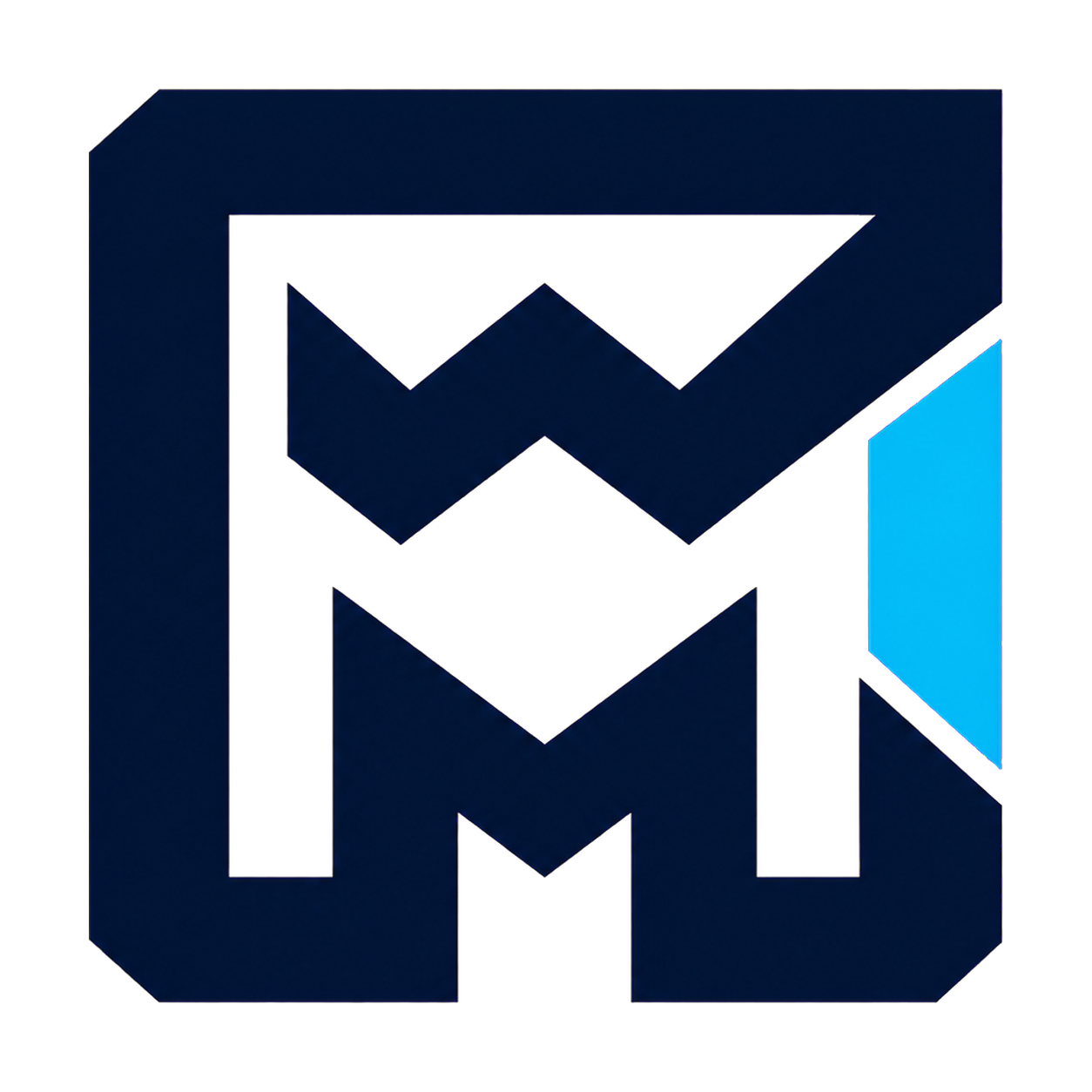}}}

\newcommand{\hflogo}{\raisebox{-0.2ex}{\includegraphics[height=2.0ex]{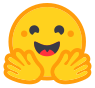}}}
\newcommand{\faGithub}{\raisebox{-0.2ex}{\includegraphics[height=2.0ex]{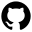}}}

\usepackage{amsmath,amsfonts,bm}

\def\eqref#1{equation~\ref{#1}}

\def\1{\bm{1}}

\def\rva{{\mathbf{a}}}

\def\vh{{\bm{h}}}

\def\vz{{\bm{z}}}

\DeclareMathAlphabet{\mathsfit}{\encodingdefault}{\sfdefault}{m}{sl}
\SetMathAlphabet{\mathsfit}{bold}{\encodingdefault}{\sfdefault}{bx}{n}

\def\sR{{\mathbb{R}}}

\newtcolorbox{promptbox}[2][]{
    colback=white,
    coltext=black,
    arc=3mm,
    boxrule=0.5pt,
    colframe=black!60!white,
    title={#2},
    colbacktitle=black,
    coltitle=white,
    fonttitle=\bfseries,
    top=8pt,
    bottom=8pt,
    left=10pt,
    right=10pt,
    breakable,
    before upper={%
        \linespread{1}\selectfont
        \setlength{\parskip}{1ex plus 0.2ex minus 0.2ex}%
        \setlength{\parindent}{0pt}%
    },
    #1
}

\title{WCM: A World Critic Model for Vision-Language-Action Reinforcement Learning}

\author{
Senyu Fei$^{1,2}$,
Xiaopeng Yu$^{3}$,
Siyin Wang$^{2,3}$,
Xianzhong Zhao$^{1}$,\\
Jingjing Gong$^{2,\dagger}$,
Xipeng Qiu$^{2,3,\dagger}$
\\[2mm] 
{\normalfont \normalsize $^{1}$Tongji University} \hspace{0.3cm}
{\normalfont \normalsize $^{2}$Shanghai Innovation Institute} \hspace{0.3cm}
{\normalfont \normalsize $^{3}$Fudan University}
\\
{\normalfont \small \texttt{feisenyu@outlook.com} \hspace{0.3cm}
\texttt{jjgongjj@gmail.com} \hspace{0.3cm}
\texttt{xpqiu@fudan.edu.cn}}
}

\abstract{
Reinforcement learning (RL) post-training of Vision-Language-Action (VLA) models has shown strong promise for robotic manipulation. Among RL methods, critic-based approaches rely on a value estimator that predominantly operates on single-frame observations or single-frame VLM backbone latents, which is a fundamental mismatch with the partially observable nature of robot control. A na\"{i}ve approach to incorporate observation history into the critic incurs exponential complexity with high-dimensional visual space, and still fails because pure scalar-return regression provides insufficient supervision for learning cross-temporal dynamics. We identify the root cause as a state approximation problem: without an explicit world modeling objective, the critic's representation cannot capture the temporal structure needed for accurate value estimation. To address this, we propose the \textbf{World Critic Model (WCM)}, built on a lightweight LeJEPA architecture; WCM jointly predicts future latent state and estimates values, such that the critic's representation is explicitly trained to capture temporal dynamics rather than merely regress scalar returns. WCM integrates seamlessly into both on-policy and off-policy training pipelines and is compatible with state-of-the-art VLA backbones including $\pi_0$, $\pi_{0.5}$, and OpenVLA-OFT. Extensive experiments on 149 tasks across four benchmarks demonstrate that WCM consistently achieves state-of-the-art performance in both in-distribution and out-of-distribution settings, with particularly strong generalization gains. We further validate WCM on seven real-world manipulation tasks using OpenVLA-OFT and $\pi_{0.5}$ with off-policy RL, confirming stable deployment across diverse settings.
}

\checkdata[\raisebox{-0.1ex}{\faGithub}\hspace{0.4em}GitHub Repo]{\url{https://github.com/sylvestf/WCM}}
\checkdata[\raisebox{-0.1ex}{\faHome}\hspace{0.4em}Homepage]{\url{https://sylvestf.github.io/wcm-homepage/}}
\checkdata[\raisebox{-0.1ex}{\hflogo}\hspace{0.4em}Hugging Face]{\url{https://huggingface.co/collections/Sylvest/wcm}}

\begin{document}
\maketitle
\begingroup
\renewcommand{\thefootnote}{\fnsymbol{footnote}}
\setcounter{footnote}{0}
\footnotetext[2]{Corresponding authors.}
\endgroup

\section{Introduction}
\label{sec:intro}

\begin{figure}[ht]
    \centering
    \includegraphics[width=\textwidth]{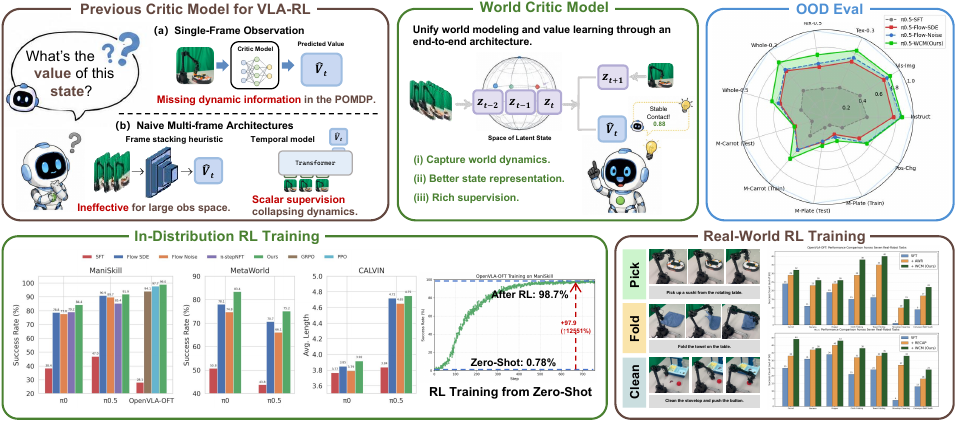}
    \caption{Overview of the World Critic Model (WCM). Prior critic models are hindered by partial observability, and na\"{i}ve multi-frame architectures remain inadequate due to sparse scalar supervision over high-dimensional observations. WCM proposes an end-to-end architecture that jointly predicts future and estimates value, effectively leveraging historical information. WCM achieves state-of-the-art performance in simulation and strong results in real-world RL training.}
    \label{fig:intro_fig}
\end{figure}

Vision-Language-Action (VLA) models have demonstrated strong performance and promising generalization in robotic manipulation~\cite{black2024pi_0,intelligence2025pi_,intelligence2026pi,kim2025fine,kim2024openvla}. Traditional supervised fine-tuning (SFT) post-training is fundamentally limited by the coverage of expert demonstrations. In contrast, reinforcement learning (RL) improves policies through environment interaction and has achieved strong performance in both in-distribution (IND) and out-of-distribution (OOD) settings~\cite{liu2025can,fei2025srpo,li2025simplevla,zang2025rlinf}. In critic-based VLA-RL methods, the critic provides dense supervision for policy improvement and is therefore a major factor in sample efficiency and final performance~\cite{intelligence2025pi,chen2025pirl,zang2026rlinf}.

Our starting point is that robotic manipulation is inherently a partially observable Markov decision process (POMDP). A single frame may reveal object appearance and scene layout, but it often misses dynamic information that is critical for value estimation, such as motion, contact progress, and possible future evolution, etc. Classical POMDP theory shows that optimal decision-making depends not on the instantaneous observation alone, but on a sufficient statistic of history and a predictive representation of state~\cite{littman2001predictive,singh2003learning,singh2012predictive,liu2022partially,subramanian2022approximate}. Therefore, a critic for VLA-RL should reason over observation history rather than a single frame alone, yet existing critics still predominantly estimate values from single-frame observations~\cite{lu2025vla,liu2025can} or VLM backbone latents~\cite{zang2025rlinf,chen2025pirl}.

However, making history available as input is not sufficient. Straightforward extensions such as frame stacking or temporal aggregation do not reliably solve the problem. Prior work shows that frame stacking can be ineffective in large observation spaces~\cite{shang2021reinforcement,efroni2022provable}, and we find that even temporal critics built with a Vision Transformer and positional encoding still struggle to improve performance. The problem is that scalar value regression provides weak supervision for learning cross-temporal dynamics: the critic can treat historical inputs as a larger static feature vector without learning how the environment evolves over time~\cite{yarats2021improving,ahuja2022weakly}.

This suggests that the missing ingredient is not history alone, but an objective that explicitly trains the critic representation to capture future evolution. The success of next-token prediction in large language model (LLM) suggests that predictive objectives can learn broadly transferable representations~\cite{achiam2023gpt,team2023gemini}. Similarly, prior work on representation learning argues that a good state representation should be able to predict its own future~\cite{schwarzer2020data}. A world-model objective is therefore appealing for VLA-RL: it provides dense supervision that complements scalar value regression and encourages the critic to encode temporally informative structure.

Motivated by the above, we propose the World Critic Model (WCM). WCM unifies world modeling and critic learning through an end-to-end architecture: given historical observations, it predicts future latent dynamics while simultaneously estimating values. Instead of treating world modeling as a detached auxiliary task, WCM uses joint optimization of predictive dynamics and value estimation to learn a critic representation that better approximates a predictive state, namely a compact and updatable summary of task-relevant future outcomes.

Extensive experiments validate the effectiveness of WCM. Across 149 tasks from four manipulation benchmarks, WCM consistently outperforms existing methods with diverse backbones, while demonstrating strong OOD generalization. On seven real-world RL tasks, it surpasses standard VLM-critic baselines, achieving better performance and smoother operation.

Our core contributions are as follows:
\begin{enumerate}
    \vspace{-5pt}
    \item We identify a representation bottleneck in critic-based VLA-RL: under partial observability, value estimation from single-frame inputs or weakly supervised history embeddings is insufficient for recovering the temporally informative state.
    \vspace{-5pt}
    \item We propose the World Critic Model (WCM), a unified critic architecture that combines future-state prediction and value estimation, so that the critic representation is explicitly trained to encode environment dynamics rather than only regress returns.
    \vspace{-5pt}
    \item We show that WCM integrates with both on-policy and off-policy VLA-RL pipelines and delivers consistent gains in performance and generalization across four simulation benchmarks, together with stable results on diverse real-world manipulation tasks.
\end{enumerate}

\vspace{-8pt}

\section{Related Works}
\label{sec:relate}

\textbf{Vision-Language-Action Reinforcement Learning (VLA-RL) has recently gained significant attention.} VLA~\cite{driess2023palm,zitkovich2023rt,kim2024openvla,black2024pi_0,pertsch2025fast,intelligence2025pi,intelligence2025pi_,intelligence2026pi} models have emerged as a promising paradigm in robot manipulation. These methods, which are pre-trained on large-scale robot manipulation and image-text data, followed by post-training on task-specific data, exhibit strong performance and promising generalization.
Representative VLAs fall into two categories: autoregressive (AR) models \cite{kim2024openvla,zitkovich2023rt,pertsch2025fast,kim2025fine} that generate actions by next token prediction, and flow-matching models \cite{intelligence2025pi,intelligence2025pi_,intelligence2026pi,black2024pi_0} that learn a continuous probability path from noise to action distribution via ordinary differential equations (ODEs).
The reliance on expert demonstrations~\cite{intelligence2025pi,fei2025srpo,zang2025rlinf,li2025simplevla} of supervised fine-tuning (SFT) has driven a growing number of studies toward RL post-training, as it offers better generalization \cite{liu2025can,li2025simplevla,fei2025srpo}, smoother deployment \cite{zang2025rlinf,chen2025pirl,intelligence2025pi}, and improved sim-to-real transfer \cite{shi2026beyond,zang2026rlinf}. For AR models, the availability of log probabilities \cite{li2025simplevla,tan2025interactive,zang2025rlinf,fei2025srpo} enables direct application of standard RL algorithms \cite{schulman2017proximal,liu2024deepseek,haarnoja2018soft,kostrikov2021offline,peng2019advantage} . Flow-matching models, by contrast, involve a deterministic ODE process and require explicit stochasticity injection for RL \cite{zhang2025reinflow,liu2025flow,chen2025pirl,ren2024diffusion}.

\textbf{Critic models in VLA-RL suffer from partial observability.} Critic model plays a central role in both on-policy \cite{wagenmaker2025steering,lu2025vla,zang2025rlinf,chen2025pirl} and off-policy \cite{luo2025precise,intelligence2025pi,peng2019advantage,team2026gigabrain} VLA-RL methods, providing dense supervision and improving sample efficiency~\cite{liu2025can,zang2025rlinf,shi2026beyond}. Existing critics typically regress values from single-frame observations \cite{intelligence2025pi,zang2025rlinf,chen2025pirl}, implicitly assuming that one frame suffices to reconstruct the system state. However, this assumption is problematic, as VLA-RL is a typical partially observable Markov decision process (POMDP)~\cite{zang2025rlinf,fei2025srpo}, which is a classical framework for decision-making under uncertainty~\cite{astrom1965optimal,smallwood1973optimal}. Relevant studies have shown that single frames may lose global or dynamic information~\cite{littman2001predictive,singh2003learning,laskin2020reinforcement,jiang2017contextual,liu2022partially}, and incorporating historical information can significantly improve policy robustness~\cite{mnih2015human,hausknecht2015deep,wang2019robust,galesloot2025robust}. Recent studies~\cite{shi2025memoryvla,koo2025hamlet,li2025cronusvla} have recognized the need for history in VLA policies, yet few works have found an effective way to incorporate history into the critic model for VLA-RL. Although incorporating history into the critic has been explored in other domains~\cite{mnih2015human,hausknecht2015deep,chen2021decision}, previous extensions fail in VLA-RL due to large observation spaces~\cite{efroni2022provable}, sparse supervision~\cite{ahuja2022weakly}, and lack of explicit learning signals for inter-frame dynamics~\cite{yarats2021improving,ahuja2022weakly}. Recently, World Action Model (WAM)~\cite{ye2026world,kim2026cosmos,bi2025motus} that jointly outputs actions and predictions demonstrates the effectiveness of world prediction~\cite{li2026causal,yuan2026fast} as a learning objective. Motivated by the above, we propose the World Critic Model (WCM), which incorporates historical information to better reconstruct system state while mitigating overfitting from single-frame value regression.

\section{Methodology}
\label{sec:method}

\subsection{Problem Formulation}

\begin{figure}[ht]
    \centering
    \includegraphics[width=\textwidth]{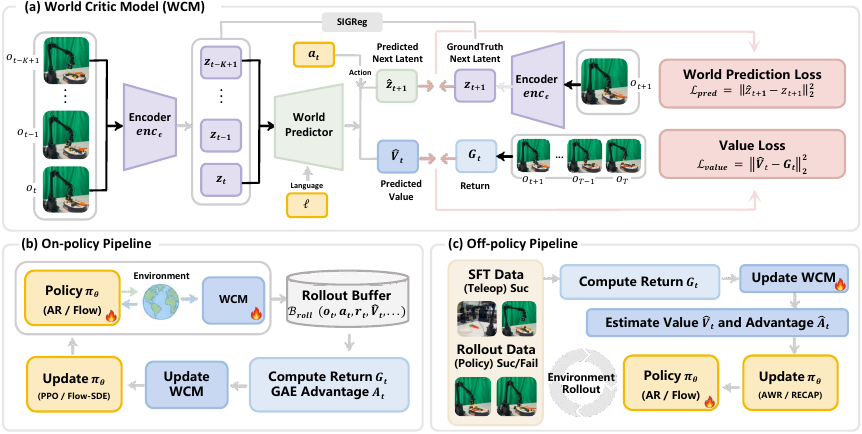}
    \caption{Overview of the World Critic Model (WCM) method. (a) \textbf{Architecture.} An observation encoder first maps individual images from the past $K$ frames into sequential latent states. Concurrently, a predictor equipped with two decoder heads forecasts the next state and estimates the value respectively. (b) \textbf{On-policy Pipeline}: WCM estimates values from on-policy rollouts to compute GAE advantages for PPO/Flow-SDE updates. (c) \textbf{Off-policy Pipeline}: Unified data buffers (SFT + rollouts) are utilized to stably update both WCM and the policy via AWR/RECAP.}
    \label{fig:wcm_diagram}
\end{figure}

\begin{wrapfigure}{r}{0.3\textwidth}
\centering
\includegraphics[width=\linewidth]{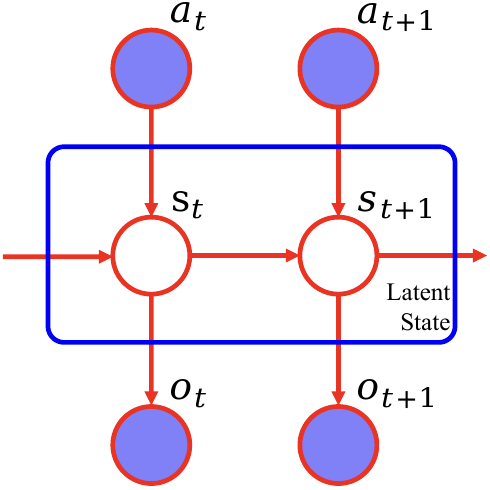}
\caption{Graphical model of the POMDP formulation for VLA-RL. The solid circles represent observable quantities, while the hollow circles represent hidden states of the system.}
\label{fig:pomdp}
\vspace{-20pt}
\end{wrapfigure}

We model robotic manipulation with VLA models as a partially observable Markov decision process (POMDP) defined by the tuple $(\mathcal{S}, \mathcal{O}, \mathcal{A}, \mathcal{T}, \mathcal{R}, \gamma)$, where $\mathcal{S}$ is the hidden state space, $\mathcal{O}$ the observation space, $\mathcal{A}$ the action space, $\mathcal{T}$ the transition function, $\mathcal{R}$ the reward function, and $\gamma$ the discount factor. As shown in Figure~\ref{fig:pomdp}, at each timestep $t$, the agent receives observation $o_t \in \mathcal{O}$ (e.g., image, proprioception, language) that partially reveals the latent state $s_t \in \mathcal{S}$, and outputs action $a_t \in \mathcal{A}$ via a VLA policy $\pi_\theta$. Then, the environment transitions from state $s_t$ to $s_{t+1}$ according to $\mathcal{T}(s_{t+1} \mid s_t, a_t)$ and the agent receives a reward $r_t = \mathcal{R}(s_t, a_t)$. The goal is to maximize $\mathbb{E}_{\pi} \left[ \sum_{t} \gamma^t r_t \right]$. 

\subsection{World Critic Model}
To unify world prediction and value estimation within a single framework while minimizing model complexity to enable training from scratch, we adopt the lightweight LeJEPA~\cite{balestriero2025lejepa,maes2026leworldmodel} architecture as the basis of our World Critic Model (WCM).

\textbf{Model Architecture.} Our model consists of four components: an observation encoder, a world predictor, and two lightweight prediction heads. Given an observation history from time steps $t-K+1$ to $t$, where $K$ is the history length, a general observation encoder first processes each observation independently into a latent embedding. Depending on the implementation, this encoder is either a ViT or the VLM backbone of the underlying VLA policy. Formally, the observation encoder produces per-frame latent embeddings as:
\begin{align}
    \vz_{t-k} = \mathrm{enc}_\epsilon(o_{t-k});~\forall k \in \{0, 1, \cdots, K-1\}.
    \label{qua1}
\end{align}
The language instruction $\ell$ is encoded by CLIP~\cite{radford2021learning}, and we map it into the WCM latent space with a learned adapter $\mathcal{A}_{\mathrm{lang}}$:
\begin{equation}
    \mathbf{u}_{\ell} =
      \mathcal{A}_{\mathrm{lang}}\!\left(
      {\mathrm{CLIP}}(\ell)
      \right)
    \in \mathbb{R}^{d}.
\end{equation}
The encoded visual history first attends to the instruction token.
The resulting language-conditioned sequence is then processed by a causal Transformer history trunk $\text{Tr}_\phi$, which we also call the world predictor. The hidden representation is given by
\begin{equation}
\vh_t = \text{Tr}_\phi(\operatorname{XAttn}
    \left(\vz_{t-K+1:t},\mathbf{u}_{\ell}\right))
\;\in\; \mathbb{R}^{d},
\end{equation}
where $d$ represents the hidden dimension. The hidden representation is fed into two separate decoder heads: a value decoder head $\mathcal{D}_{\text{value}}$ for return estimation. The value estimation is given by
\begin{equation}
\hat{V}_t = \mathcal{D}_{\text{value}}(\vh_t)
\;\in\; \mathbb{R},
\label{qua3}
\end{equation}
and the action-conditioned latent dynamics branch predicts the next latent using a residual update:
\begin{equation}
    \hat{\vz}_{t+1}
    =
    \mathcal{D}_{\text{world}}(\vh_t,a_t,\vz_t)
    \in \mathbb{R}^{d},
\end{equation}
where $\mathcal{D}_{\text{world}}$ is implemented with an action encoder and gated
FiLM~\cite{perez2018film} residual blocks.

\textbf{Training Objective.} The overall training objective combines three components: the prediction loss, the Sketched-Isotropic Gaussian Regularization (SIGReg)~\cite{balestriero2025lejepa} loss, and the value estimation loss.
The \textbf{prediction loss} $ \mathcal{L}_{\text{pred}} $ uses teacher-forcing to compute the error between the predicted next latent state and the true next latent state:
\begin{equation}
\mathcal{L}_{\text{pred}} = \| \hat{\vz}_{t+1} - \vz_{t+1} \|_2^2,
\label{eq:4}
\end{equation}
where $ \hat{\vz}_{t+1} $ is the predicted latent state and $ \vz_{t+1} $ is the ground truth.

To prevent feature collapse in the latent space, we apply Sketched-Isotropic Gaussian Regularization (SIGReg)~\cite{balestriero2025lejepa} to the latent representations $\vz_t$, encouraging them to match an isotropic Gaussian distribution. The \textbf{SIGReg loss} is defined as:
\begin{equation}
    \mathcal{L}_{\text{SIGReg}} = \mathbb{E}_{\rva \sim \mathcal{U}(\mathcal{S}^{d-1})} \left[ \int_{\sR} \left|\hat{\phi}_{\rva^\top \vz}(t) - \phi(t)\right|^2 e^{-t^2}\, dt \right],
\end{equation}
where $\rva$ is a random unit vector drawn uniformly from the $(d{-}1)$-sphere, $\phi(t) = e^{-t^2/2}$ is the characteristic function of a standard normal, and $\hat{\phi}_{\rva^\top \vz}(t) = \mathbb{E}_{\vz}\left[e^{it(\rva^\top \vz)}\right]$ is the empirical characteristic function of the projected representation, approximated over a mini-batch. Intuitively, SIGReg enforces that every one-dimensional projection of $\vz$ onto a random unit vector $\rva$ matches the characteristic function of a standard Gaussian, which is a necessary and sufficient condition for $\vz$ to follow an isotropic Gaussian distribution. This penalizes dimensional collapse and mode degeneration in the learned latent space.

The \textbf{value loss} $ \mathcal{L}_{\text{value}} $ aims to predict the return associated with each state. Building on the insights from prior work~\cite{intelligence2025pi}, for each time step in the trajectory, the reward $ r_{t} $ and return $ G_{t} $ are given as:
\begin{equation}
r_t = \begin{cases} 
0 & \text{if \(t=T\) and success,} \\
-C_{\text{fail}} & \text{if \(t=T\) and failure,} \\
-1 & \text{otherwise,}
\end{cases}
\ \ \ 
G_t = \sum_{t' = t}^{T} \gamma^{t' - t} r_{t'},
\label{qua5}
\end{equation}
where $T$ is the last step in the episode, $C_{\text{fail}}$ is a large positive constant to penalize poor performance, and $ \gamma $ is the discount factor.
Then we min-max normalize the returns to be in $[-1, 1]$.
The value loss $ \mathcal{L}_{\text{value}} $ is the L2 loss between the predicted value $ \hat{V}_t $ and the ground truth return $ G_t $:
\begin{equation}
\mathcal{L}_{\text{value}} = \| \hat{V}_t - G_t \|_2^2.
\label{eq:7}
\end{equation}
The \textbf{complete training objective} is then:
\begin{equation}
\mathcal{L} = \mathcal{L}_{\text{value}} + \lambda \cdot \mathcal{L}_{\text{pred}} + \eta \cdot \mathcal{L}_{\text{SIGReg}},
\label{eq:9}
\end{equation}
where $ \lambda $ is a hyperparameter controlling the weight of the prediction loss, and $ \eta $ is a hyperparameter controlling the weight of the SIGReg regularization. All components are trained end-to-end.

\subsection{Training Pipeline}

\textbf{On-policy Setting:} For Auto-Regressive (AR) models such as OpenVLA-OFT~\cite{kim2025fine}, we employ the Proximal Policy Optimization (PPO)~\cite{schulman2017proximal} algorithm. For flow-matching models like $\pi_{0}$~\cite{black2024pi_0} and $\pi_{0.5}$~\cite{intelligence2025pi_}, we utilize Flow-SDE~\cite{chen2025pirl}, a variant of PPO. In both frameworks, we adopt WCM as the critic model. To maintain a lean architecture and leverage the representations already learned during pretraining, we use the VLM backbone as the observation encoder in WCM. The detailed algorithm is outlined in Appendix~\ref{app:appendix_algorithms} Algorithm~\ref{algo:ppo_wcm_update}.

\textbf{Off-policy Setting:} In the off-policy setting, each training iteration incorporates not only SFT data collected via teleoperation but also data from erroneous rollouts and typical failure cases. (Rollout data is unavailable in the first iteration.) These additional data sources improve the accuracy of value estimation and help the WCM learn more realistic predictions, avoiding overly optimistic estimates. For AR models, we adopt the Advantage-weighted Regression (AWR)~\cite{peng2019advantage}. For flow-matching models, we use the RECAP~\cite{intelligence2025pi} method introduced by $\pi^{*}_{0.6}$. In both cases, we instantiate the critic model with WCM. The algorithms are shown in Appendix~\ref{app:appendix_algorithms} Algorithms~\ref{algo:awr} and~\ref{algo:recap}.

\section{Experiments}
\label{sec:exp}
\textbf{TL;DR;} We carefully designed our experiments and arrived at the following conclusions. (1) WCM consistently improves performance across simulation manipulation benchmarks. (2) WCM exhibits stronger generalization to OOD settings compared with existing methods. (3) WCM performs effectively in real-world RL training. (4) The world prediction objective plays a positive role in leveraging historical information. (5) Longer state history provides limited benefits beyond a certain optimal length rather than universal improvement in our tasks.

\subsection{Experimental Setup}
\textbf{Simulation.}  
We evaluate our method on 149 tasks across four simulation benchmarks. Following the RL4VLA~\cite{liu2025can} setup, we leverage ManiSkill~\cite{mu2021maniskill} and assess both in-distribution (IND) and out-of-distribution (OOD) performance across three axes.
Additionally, we evaluate on MetaWorld~\cite{yu2020meta} for performance on tasks beyond pick-and-place, CALVIN~\cite{mees2022calvin} for long-horizon capabilities, and LIBERO-Plus~\cite{fei2025libero} for generalization abilities. All policies are initialized from few-shot SFT baselines and trained using sparse 0/1 rewards, including $\pi_0$~\cite{black2024pi_0}, $\pi_{0.5}$~\cite{intelligence2025pi_}, and OpenVLA-OFT~\cite{kim2025fine}.

\textbf{Real-world.} We evaluate our method on 7 robotic tasks on WidowX-250S: a dynamic grasping task, a long-horizon task, 2 deformable object manipulation tasks, and 3 pick-and-place tasks. Using $\pi_{0.5}$ and OpenVLA-OFT as base policies, we train them with off-policy RL guided by our WCM.

\textbf{Baselines.} For simulation, we adopt on-policy RL methods as baselines. To be specific, for $\pi$-style policies, we consider three state-of-the-art baselines: Flow-Noise~\cite{zhang2025reinflow}, Flow-SDE~\cite{ren2024diffusion}, and $\pi$-stepNFT~\cite{wang2026pi}. 
For OpenVLA-OFT, we leverage the RLinf~\cite{chen2025pirl} implementations of standard PPO~\cite{schulman2017proximal} and GRPO~\cite{liu2024deepseek} as baselines. 
For real-robot, we adopt off-policy methods as baselines, specifically, AWR~\cite{peng2019advantage} for OpenVLA-OFT and RECAP~\cite{intelligence2025pi} for $\pi_{0.5}$. More details in Appendix~\ref{app:appendix_algorithms}.

\subsection{Main Results}

\begin{table}[t]
\centering
\small
\caption{
Performance comparison on ManiSkill under IND and OOD settings.
All methods are trained via RL on top of SFT initialization, with $\Delta$ denotes the performance improvement relative to the SFT baseline. ``+WCM'' denotes replacing the critic model with WCM on Flow-SDE or PPO.
}
\setlength{\tabcolsep}{2.8pt}
\renewcommand{\arraystretch}{1.2}
\begin{tabular}{l l cc|ccc|cc}
\toprule
\multirow{2}{*}{Backbone} & \multirow{2}{*}{Method} 
& \multicolumn{2}{c}{IND} 
& \multicolumn{5}{c}{OOD} \\
\cmidrule(lr){3-4} \cmidrule(lr){5-9}
& 
& avg. & $\Delta$ 
& vision & semantic & execution 
& avg. & $\Delta$ \\
\midrule

\multirow{5}{*}{$\pi_0$}
& SFT & 38.4 & - & 32.6 & 8.4 & 13.2 & 18.1 & - \\
& + FlowSDE~\cite{chen2025pirl,ren2024diffusion} & 78.8 & +40.4 & 61.1 & 25.4 & 31.5 & 39.3 & +21.2 \\
& + FlowNoise~\cite{chen2025pirl,zhang2025reinflow} & 77.8 & +39.4 & 63.4 & 23.1 & 24.2 & 36.9 & +18.8 \\
& + $\pi$-stepNFT~\cite{wang2026pi} & 79.2 & +40.8 & \textbf{69.1} & 49.1 & 33.1 & 50.4 & +32.3 \\
\rowcolor{gray!10}
& + WCM (Ours) & \textbf{84.4}\scriptsize{$\pm$1.2} & \textbf{+46.0}\scriptsize{$\pm$1.2} & \textbf{69.1}\scriptsize{$\pm$0.7} & \textbf{49.8}\scriptsize{$\pm$2.2} & \textbf{35.6}\scriptsize{$\pm$1.3} & \textbf{51.5}\scriptsize{$\pm$1.5} & \textbf{+33.4}\scriptsize{$\pm$1.5} \\

\midrule

\multirow{5}{*}{$\pi_{0.5}$}
& SFT & 47.0 & - & 40.2 & 16.6 & 22.4 & 26.4 & - \\
& + FlowSDE~\cite{chen2025pirl,ren2024diffusion} & 90.9 & +43.9 & 68.0 & 34.5 & 45.4 & 49.3 & +22.9 \\
& + FlowNoise~\cite{chen2025pirl,zhang2025reinflow} & 89.7 & +42.7 & 69.9 & 35.5 & 54.9 & 53.4 & +27.0 \\
& + $\pi$-stepNFT~\cite{wang2026pi} & 85.4 & +38.4 & 76.9 & 56.6 & 45.1 & 59.5 & +33.1 \\
\rowcolor{gray!10}
& + WCM (Ours) & \textbf{91.9}\scriptsize{$\pm$0.4} & \textbf{+44.9}\scriptsize{$\pm$0.4} & \textbf{78.1}\scriptsize{$\pm$1.6} & \textbf{58.5}\scriptsize{$\pm$1.0} & \textbf{56.5}\scriptsize{$\pm$1.5} & \textbf{64.4}\scriptsize{$\pm$1.4} & \textbf{+38.0}\scriptsize{$\pm$1.4} \\

\midrule

\multirow{4}{*}{\makecell[l]{OpenVLA-OFT}}
& SFT & 28.1 & - & 27.7 & 13.0 & 11.7 & 18.3 & - \\
& + GRPO~\cite{zang2025rlinf,liu2024deepseek} & 94.1 & +66.0 & 84.7 & 45.5 & 44.7 & 60.6 & +42.3 \\
& + PPO~\cite{zang2025rlinf,schulman2017proximal} & 97.7 & +69.6 & 92.1 & 64.8 & 73.6 & 77.1 & +58.8 \\
\rowcolor{gray!10}
& + WCM (Ours) & \textbf{99.0}\scriptsize{$\pm$0.4} & \textbf{+70.9}\scriptsize{$\pm$0.4} & \textbf{92.4}\scriptsize{$\pm$0.5} & \textbf{65.9}\scriptsize{$\pm$1.0} & \textbf{75.5}\scriptsize{$\pm$0.7} & \textbf{77.9}\scriptsize{$\pm$0.8} & \textbf{+59.6}\scriptsize{$\pm$0.8} \\
& Zero-Shot & 0.8 & - & 0.5 & 1.0 & 1.0 & 0.8 & - \\
\rowcolor{gray!10}
& + WCM (Ours) & 98.7\scriptsize{$\pm$0.3} & +97.9\scriptsize{$\pm$0.3} & 88.0\scriptsize{$\pm$1.2} & 62.4\scriptsize{$\pm$2.9} & 70.1\scriptsize{$\pm$0.3} & 73.5\scriptsize{$\pm$1.8} & +72.7\scriptsize{$\pm$1.8} \\

\bottomrule
\end{tabular}
\label{tab:main_results}
\end{table}

\begin{wrapfigure}{r}{0.5\textwidth}
\centering
\includegraphics[width=\linewidth]{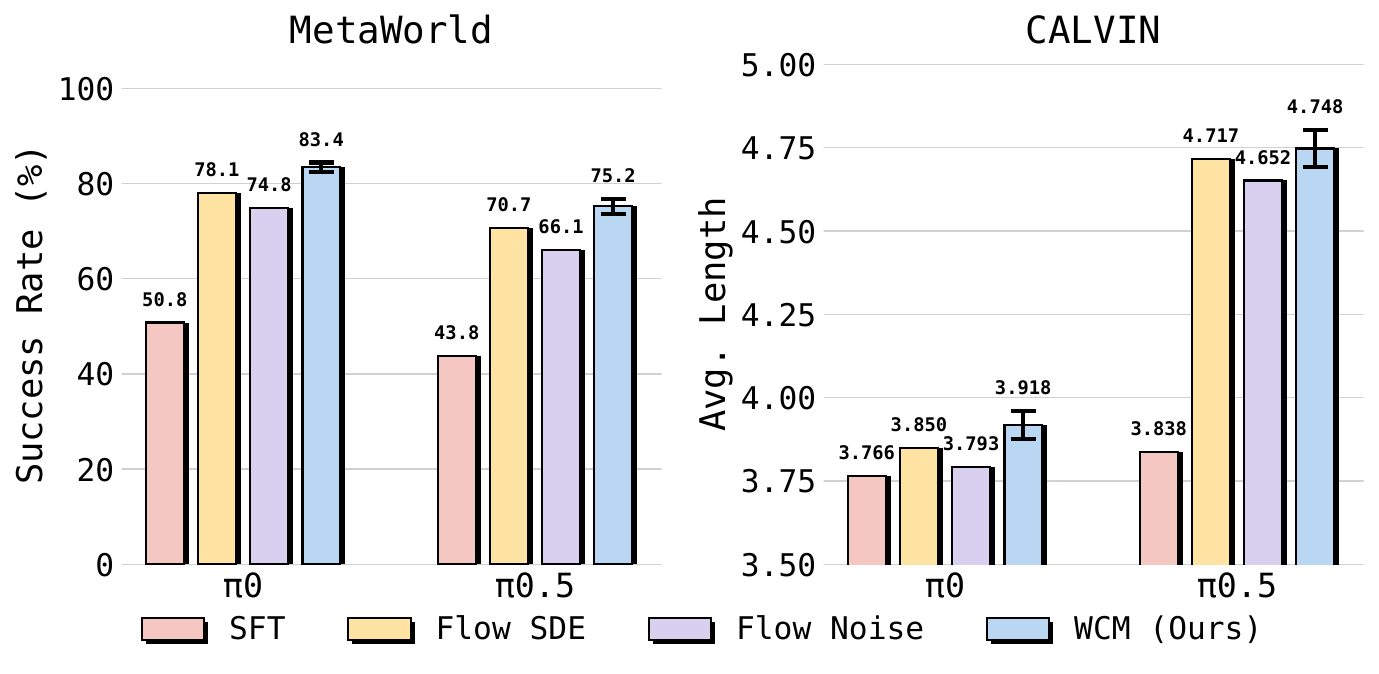}
\caption{Results on MetaWorld and CALVIN. We report success rate or average length with error bars.}
\label{fig:mw_calvin}
\vspace{-20pt}
\end{wrapfigure}

The main results are presented in Table~\ref{tab:main_results} and Figure~\ref{fig:mw_calvin}. 
Our method achieves SOTA performance across ManiSkill, MetaWorld and CALVIN benchmarks. 
Key findings are as follows:

\textbf{(1) Significant performance gains on ManiSkill.}  
Our method significantly enhances performance from weak SFT baselines, especially for OpenVLA-OFT with a 252\% improvement.

\textbf{(2) Stable improvement from extremely low initial performance.}  
When OpenVLA-OFT is initialized with no exposure to ManiSkill data (0.78\%), our method improves performance by 12,551\%.

\textbf{(3) Effectiveness on diverse tasks and long-horizon scenarios.} 
On MetaWorld, it outperforms in tasks requiring stable contact beyond pick-and-place tasks. On CALVIN, improvement reflects stronger long-horizon capabilities. These gains stem from our method’s ability to capture higher-order state information, enhancing latent representation and value estimation in POMDP settings.

\begin{table*}[t]
\centering
\small
\caption{Detailed performance on LIBERO-Plus. ``One-SFT'' is the SFT model seeing only one demonstration per task, while ``Full-SFT'' seeing 50. ``+WCM'' is initialized from One-SFT.}
\setlength{\tabcolsep}{2.8pt}
\renewcommand{\arraystretch}{1.2}
\begin{tabular}{l l cccccccc}
\toprule
\multirow{2}{*}{Backbone} & \multirow{2}{*}{Method}
& \multicolumn{8}{c}{Metrics} \\
\cmidrule(lr){3-10}
& & camera & env & init & language & noise & layout & light & total \\
\midrule

\multirow{4}{*}{$\pi_0$}
& Full-SFT & 85.4\scriptsize{$\pm$1.4} & 90.1\scriptsize{$\pm$1.8} & 12.5\scriptsize{$\pm$0.9} & 65.8\scriptsize{$\pm$1.3} & \textbf{89.7}\scriptsize{$\pm$2.5} & 75.3\scriptsize{$\pm$1.2} & 89.3\scriptsize{$\pm$0.7} & 71.2\scriptsize{$\pm$1.5} \\
& One-SFT & 32.3\scriptsize{$\pm$2.2} & 59.4\scriptsize{$\pm$0.2} & 1.5\scriptsize{$\pm$0.9} & 42.3\scriptsize{$\pm$1.7} & 48.7\scriptsize{$\pm$0.4} & 50.2\scriptsize{$\pm$2.8} & 48.0\scriptsize{$\pm$3.8} & 39.1\scriptsize{$\pm$2.1} \\
& + WCM (Ours) & \textbf{88.3}\scriptsize{$\pm$2.7} & \textbf{91.0}\scriptsize{$\pm$1.8} & \textbf{19.3}\scriptsize{$\pm$1.0} & \textbf{65.9}\scriptsize{$\pm$0.2} & 87.5\scriptsize{$\pm$0.2} & \textbf{76.0}\scriptsize{$\pm$1.4} & \textbf{90.9}\scriptsize{$\pm$3.6} & \textbf{72.8}\scriptsize{$\pm$1.9} \\
\rowcolor{gray!10}
& $\Delta$
& +56.0 & +31.6 & +17.8 & +23.6 & +38.8 & +25.8 & +42.9 & +33.7 \\

\midrule

\multirow{4}{*}{$\pi_{0.5}$}
& Full-SFT & 78.8\scriptsize{$\pm$0.5} & 89.9\scriptsize{$\pm$2.9} & 24.7\scriptsize{$\pm$1.2} & 73.8\scriptsize{$\pm$1.0} & \textbf{88.2}\scriptsize{$\pm$1.7} & 77.0\scriptsize{$\pm$2.0} & 78.3\scriptsize{$\pm$1.9} & 72.9\scriptsize{$\pm$1.8} \\
& One-SFT & 32.8\scriptsize{$\pm$0.5} & 52.7\scriptsize{$\pm$3.0} & 1.9\scriptsize{$\pm$0.6} & 38.7\scriptsize{$\pm$0.7} & 50.5\scriptsize{$\pm$2.2} & 46.1\scriptsize{$\pm$1.5} & 51.2\scriptsize{$\pm$0.6} & 38.0\scriptsize{$\pm$1.6} \\
& + WCM (Ours) & \textbf{80.8}\scriptsize{$\pm$1.7} & \textbf{90.3}\scriptsize{$\pm$1.6} & \textbf{31.5}\scriptsize{$\pm$0.6} & \textbf{65.0}\scriptsize{$\pm$1.1} & 86.3\scriptsize{$\pm$2.1} & \textbf{79.3}\scriptsize{$\pm$0.9} & \textbf{91.6}\scriptsize{$\pm$0.8} & \textbf{73.7}\scriptsize{$\pm$1.4} \\
\rowcolor{gray!10}
& $\Delta$
& +48.0 & +37.6 & +29.6 & +26.3 & +35.8 & +33.2 & +40.4 & +35.7 \\

\midrule

\multirow{4}{*}{\makecell[l]{OpenVLA-OFT}}
& Full-SFT & 69.4\scriptsize{$\pm$1.5} & 88.5\scriptsize{$\pm$0.6} & 49.6\scriptsize{$\pm$1.2} & \textbf{66.3}\scriptsize{$\pm$1.1} & 78.7\scriptsize{$\pm$1.3} & \textbf{70.3}\scriptsize{$\pm$1.0} & 88.2\scriptsize{$\pm$2.0} & 71.7\scriptsize{$\pm$1.3} \\
& One-SFT & 12.8\scriptsize{$\pm$0.8} & 49.6\scriptsize{$\pm$1.9} & 23.0\scriptsize{$\pm$1.0} & 30.0\scriptsize{$\pm$2.2} & 23.3\scriptsize{$\pm$2.3} & 34.5\scriptsize{$\pm$0.3} & 42.0\scriptsize{$\pm$0.1} & 29.3\scriptsize{$\pm$1.5} \\
& + WCM (Ours) & \textbf{74.6}\scriptsize{$\pm$3.6} & \textbf{94.9}\scriptsize{$\pm$0.1} & \textbf{51.3}\scriptsize{$\pm$0.4} & 65.8\scriptsize{$\pm$2.2} & \textbf{84.1}\scriptsize{$\pm$0.6} & 63.6\scriptsize{$\pm$1.4} & \textbf{94.8}\scriptsize{$\pm$1.6} & \textbf{74.0}\scriptsize{$\pm$1.8} \\
\rowcolor{gray!10}
& $\Delta$
& +61.8 & +45.3 & +28.3 & +35.8 & +60.8 & +29.1 & +52.8 & +44.7 \\

\bottomrule
\end{tabular}
\label{tab:detailed_results}
\end{table*}

\subsection{Generalization Performance}

Generalization is a key evaluation criterion in VLA-RL. 
We evaluate our method under ManiSkill-OOD setting and LIBERO-Plus. 
Results are reported in Table~\ref{tab:main_results} and Table~\ref{tab:detailed_results}. Key findings:

\textbf{(1) Strong generalization gains with WCM.}  
On ManiSkill, WCM improves both IND performance and OOD generalization, outperforming traditional Flow-SDE and PPO. Additionally, our method outperforms $\pi$-StepNFT baseline, which is known for its strong OOD performance, benefiting from WCM that captures more state information and robust value estimation for distribution shifts.

\textbf{(2) Superior generalization than SFT.}  
In LIBERO-Plus, starting from one-shot SFT, after about 250 RL training steps, our method outperforms full-shot SFT trained on 20k trajectories.

\subsection{Real-World Performance}

In the real-world setting, for better sample efficiency, we employ the off-policy pipeline. We select two RL algorithms as baselines: AWR for OpenVLA-OFT and RECAP for $\pi_{0.5}$, for which we use Gemma 270M as the critic model.
We evaluate all methods on 7 tasks: dynamic manipulation (rotating sushi picking), deformable object manipulation (cloth \& towel folding), long-horizon (stovetop cleaning), and pick-and-place (carrot, pepper, banana) tasks. We train the SFT policy using 100 trajectories per task and perform 8 RL iterations, with 50 rollouts per task per iteration. As shown in Table~\ref{tab:real_robot_results}, our method (WCM with 107.2M learnable parameters) outperforms baselines across all tasks, attributed to its better state reconstruction and consequently more accurate value estimation.

It is noteworthy that the real-world experiments demonstrate the effectiveness and efficiency of WCM under limited-data training on physical robots, operating at the scale of hundreds to a few thousand data samples. With only hundreds to a few thousand trajectories and less than one hour of training, WCM enables rapid iterative refinement and yields accurate critic predictions.
Corresponding dynamic visualizations are available in our \href{https://github.com/sylvestf/WCM}{repository} and on our \href{https://sylvestf.github.io/wcm-homepage/}{project website}.

\section{Analysis}
\label{sec:ana}

\begin{table*}[t]
\centering
\small
\caption{Detailed real-world experiment results. ``+WCM'' denotes using WCM as critic model. $\Delta$ denotes improvement from SFT. Results are taken from the first 50 trajectories after test starts.}
\setlength{\tabcolsep}{2.8pt}
\renewcommand{\arraystretch}{1.2}
\begin{tabular}{l l ccc|cc|c|cc}
\toprule
\multirow{3}{*}{Backbone} & \multirow{3}{*}{Method}
& \multicolumn{3}{c}{Pick and Place} & \multicolumn{2}{c}{Deformable} & \multicolumn{1}{c}{Long-Horizon} & \multicolumn{1}{c}{Moving} \\
\cmidrule(lr){3-9}
& & Carrot & Banana & Pepper & \multicolumn{1}{c}{\begin{tabular}[c]{@{}c@{}}Cloth\\Folding\end{tabular}} & \multicolumn{1}{c|}{\begin{tabular}[c]{@{}c@{}}Towel\\Folding\end{tabular}} & \multicolumn{1}{c|}{\begin{tabular}[c]{@{}c@{}}Stovetop\\Cleaning\end{tabular}} & \multicolumn{1}{c}{\begin{tabular}[c]{@{}c@{}}Conveyor Belt\\ Sushi Picking\end{tabular}} \\
\midrule
\multirow{4}{*}{OpenVLA-OFT}
& SFT & 24/50 & 11/50 & 19/50 & 15/50 & 16/50 & 1/50 & 9/50 \\
& + AWR~\cite{peng2019advantage} & 29/50 & 23/50 & 24/50 & 29/50 & 35/50 & 10/50 & 17/50 \\
& + WCM (Ours) & 32/50 & 26/50 & 26/50 & 38/50 & 40/50 & 15/50 & 22/50 \\
\rowcolor{gray!10}
& $\Delta$ & 8/50 & 15/50 & 7/50 & 23/50 & 24/50 & 14/50 & 13/50 \\
\midrule
\multirow{4}{*}{$\pi_{0.5}$}
& SFT & 25/50 & 31/50 & 34/50 & 21/50 & 24/50 & 4/50 & 13/50 \\
& + RECAP~\cite{intelligence2025pi} & 33/50 & 37/50 & 40/50 & 32/50 & 33/50 & 27/50 & 18/50 \\
& + WCM (Ours) & 44/50 & 38/50 & 43/50 & 38/50 & 35/50 & 33/50 & 24/50 \\
\rowcolor{gray!10}
& $\Delta$ & 19/50 & 7/50 & 9/50 & 17/50 & 11/50 & 29/50 & 11/50 \\
\bottomrule
\end{tabular}
\label{tab:real_robot_results}
\end{table*}

\subsection{Does World Prediction Objective Matter?}
\label{exp:world-predict}
\begin{figure}[htbp]
    \centering
    \includegraphics[width=0.9\textwidth]{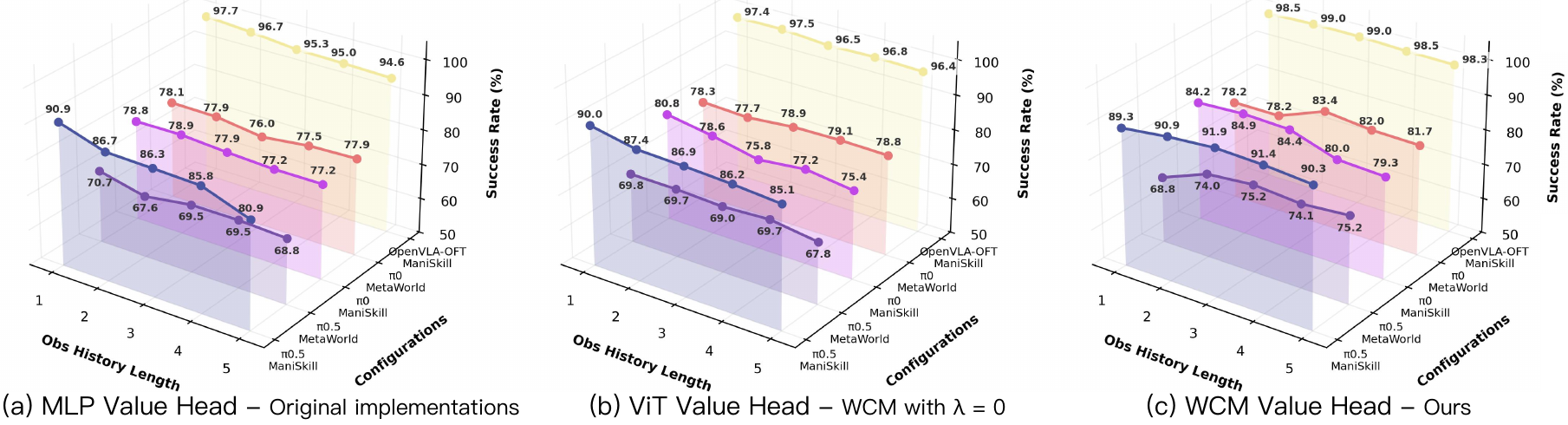}
    \caption{Ablation study on critic architectures and observation history lengths. WCM achieves the highest success rates across all configurations. MLP denotes the baseline critic in PPO/Flow-SDE; ViT represents WCM without world prediction ($\lambda = 0$), which has temporal modeling capability.}
    \label{fig:history_ablation}
\end{figure}

We further validate the necessity of the world prediction objective through additional experiments, as shown in Figure~\ref{fig:history_ablation}. We evaluate three models ($\pi_0$, $\pi_{0.5}$, and OpenVLA-OFT) on ManiSkill and MetaWorld, extending the critic's input from single-frame to 2-5 frames. Key findings are as follows:
\textbf{(1)} Modifying the original MLP critic model by incorporating more observation history may lead to suboptimal performance.
\textbf{(2)} Leveraging a history-based ViT (a special case of WCM with $\lambda=0$) still proves ineffective.
\textbf{(3)} Incorporating a world prediction objective enables better performance.

\subsection{Is longer observation history always beneficial?}
\label{exp:longer}

We conduct an ablation study by varying the observation history length of WCM from 1 to 5, and report the best IND performance in Figure~\ref{fig:history_ablation}. In our experiments, length 3 performed best on average. A plausible intuitive explanation is that three consecutive frames may implicitly capture second-order dynamics (acceleration), while two frames capture first-order dynamics (velocity). For our tasks, first- and second-order information seem sufficient to describe the required dynamic features.

\section{Conclusion}
\label{sec:con}

In this work, we identified a fundamental limitation of existing critic-based VLA-RL methods: value estimation from single-frame observations or weakly supervised history fails to capture the temporal structure required for state reconstruction under partial observability. To address this, we proposed the World Critic Model (WCM), a unified architecture that jointly learns latent state prediction and value estimation. Extensive experiments on 149 tasks across four simulation benchmarks demonstrate that WCM consistently achieves state-of-the-art performance in both in-distribution and out-of-distribution settings, with particularly strong generalization gains. We further validate WCM on seven real-world manipulation tasks, confirming its stable and effective deployment.

\clearpage
\bibliographystyle{unsrtnat}
\bibliography{main}

@article{liu2025can,
  title={What can rl bring to vla generalization? an empirical study},
  author={Liu, Jijia and Gao, Feng and Wei, Bingwen and Chen, Xinlei and Liao, Qingmin and Wu, Yi and Yu, Chao and Wang, Yu},
  journal={arXiv preprint arXiv:2505.19789},
  year={2025}
}

@article{fei2025libero,
  title={Libero-plus: In-depth robustness analysis of vision-language-action models},
  author={Fei, Senyu and Wang, Siyin and Shi, Junhao and Dai, Zihao and Cai, Jikun and Qian, Pengfang and Ji, Li and He, Xinzhe and Zhang, Shiduo and Fei, Zhaoye and others},
  journal={arXiv preprint arXiv:2510.13626},
  year={2025}
}

@article{intelligence2025pi,
  title={$\pi_{0.6}$: A VLA That Learns From Experience},
  author={Intelligence, Physical and Amin, Ali and Aniceto, Raichelle and Balakrishna, Ashwin and Black, Kevin and Conley, Ken and Connors, Grace and Darpinian, James and Dhabalia, Karan and DiCarlo, Jared and others},
  journal={arXiv preprint arXiv:2511.14759},
  year={2025}
}

@article{kim2025fine,
  title={Fine-tuning vision-language-action models: Optimizing speed and success},
  author={Kim, Moo Jin and Finn, Chelsea and Liang, Percy},
  journal={arXiv preprint arXiv:2502.19645},
  year={2025}
}

@article{fei2025srpo,
  title={SRPO: Self-Referential Policy Optimization for Vision-Language-Action Models},
  author={Fei, Senyu and Wang, Siyin and Ji, Li and Li, Ao and Zhang, Shiduo and Liu, Liming and Hou, Jinlong and Gong, Jingjing and Zhao, Xianzhong and Qiu, Xipeng},
  journal={arXiv preprint arXiv:2511.15605},
  year={2025}
}

@article{peng2019advantage,
  title={Advantage-weighted regression: Simple and scalable off-policy reinforcement learning},
  author={Peng, Xue Bin and Kumar, Aviral and Zhang, Grace and Levine, Sergey},
  journal={arXiv preprint arXiv:1910.00177},
  year={2019}
}

@article{mu2021maniskill,
  title={Maniskill: Generalizable manipulation skill benchmark with large-scale demonstrations},
  author={Mu, Tongzhou and Ling, Zhan and Xiang, Fanbo and Yang, Derek and Li, Xuanlin and Tao, Stone and Huang, Zhiao and Jia, Zhiwei and Su, Hao},
  journal={arXiv preprint arXiv:2107.14483},
  year={2021}
}

@article{mees2022calvin,
  title={Calvin: A benchmark for language-conditioned policy learning for long-horizon robot manipulation tasks},
  author={Mees, Oier and Hermann, Lukas and Rosete-Beas, Erick and Burgard, Wolfram},
  journal={IEEE Robotics and Automation Letters},
  volume={7},
  number={3},
  pages={7327--7334},
  year={2022},
  publisher={IEEE}
}

@inproceedings{yu2020meta,
  title={Meta-world: A benchmark and evaluation for multi-task and meta reinforcement learning},
  author={Yu, Tianhe and Quillen, Deirdre and He, Zhanpeng and Julian, Ryan and Hausman, Karol and Finn, Chelsea and Levine, Sergey},
  booktitle={Conference on robot learning},
  pages={1094--1100},
  year={2020},
  organization={PMLR}
}

@article{black2024pi_0,
  title={$\pi_{0}$: A Vision-Language-Action Flow Model for General Robot Control},
  author={Black, Kevin and Brown, Noah and Driess, Danny and Esmail, Adnan and Equi, Michael and Finn, Chelsea and Fusai, Niccolo and Groom, Lachy and Hausman, Karol and Ichter, Brian and others},
  journal={arXiv preprint arXiv:2410.24164},
  year={2024}
}

@article{intelligence2025pi_,
  title={$\pi_{0.5}$: A Vision-Language-Action Model with Open-World Generalization},
  author={Intelligence, Physical and Black, Kevin and Brown, Noah and Darpinian, James and Dhabalia, Karan and Driess, Danny and Esmail, Adnan and Equi, Michael and Finn, Chelsea and Fusai, Niccolo and others},
  journal={9th Annual Conference on Robot Learning},
  year={2025}
}

@article{chen2025pirl,
  title={$\pi$RL: Online RL Fine-Tuning for Flow-Based Vision-Language-Action Models},
  author={Chen, Kang and Liu, Zhihao and Zhang, Tonghe and Guo, Zhen and Xu, Si and Lin, Hao and Zang, Hongzhi and Zhang, Quanlu and Yu, Zhaofei and Fan, Guoliang and others},
  journal={arXiv preprint arXiv:2510.25889},
  year={2025}
}

@article{schulman2017proximal,
  title={Proximal policy optimization algorithms},
  author={Schulman, John and Wolski, Filip and Dhariwal, Prafulla and Radford, Alec and Klimov, Oleg},
  journal={arXiv preprint arXiv:1707.06347},
  year={2017}
}

@article{liu2024deepseek,
  title={Deepseek-v3 technical report},
  author={Liu, Aixin and Feng, Bei and Xue, Bing and Wang, Bingxuan and Wu, Bochao and Lu, Chengda and Zhao, Chenggang and Deng, Chengqi and Zhang, Chenyu and Ruan, Chong and others},
  journal={arXiv preprint arXiv:2412.19437},
  year={2024}
}

@article{wang2026pi,
  title={$\pi$-StepNFT: Wider Space Needs Finer Steps in Online RL for Flow-Based VLAs},
  author={Wang, Siting and Wang, Xiaofeng and Zhu, Zheng and Pei, Minnan and Cui, Xinyu and Deng, Cheng and Zhao, Jian and Huang, Guan and Zhang, Haifeng and Wang, Jun},
  journal={arXiv preprint arXiv:2603.02083},
  year={2026}
}

@article{zhang2025reinflow,
  title={ReinFlow: Fine-tuning flow matching policy with online reinforcement learning},
  author={Zhang, Tonghe and Yu, Chao and Su, Sichang and Wang, Yu},
  journal={arXiv preprint arXiv:2505.22094},
  year={2025}
}

@article{ren2024diffusion,
  title={Diffusion policy policy optimization},
  author={Ren, Allen Z and Lidard, Justin and Ankile, Lars L and Simeonov, Anthony and Agrawal, Pulkit and Majumdar, Anirudha and Burchfiel, Benjamin and Dai, Hongkai and Simchowitz, Max},
  journal={arXiv preprint arXiv:2409.00588},
  year={2024}
}

@article{kim2024openvla,
  title={Openvla: An open-source vision-language-action model},
  author={Kim, Moo Jin and Pertsch, Karl and Karamcheti, Siddharth and Xiao, Ted and Balakrishna, Ashwin and Nair, Suraj and Rafailov, Rafael and Foster, Ethan and Lam, Grace and Sanketi, Pannag and others},
  journal={arXiv preprint arXiv:2406.09246},
  year={2024}
}

@inproceedings{zitkovich2023rt,
  title={Rt-2: Vision-language-action models transfer web knowledge to robotic control},
  author={Zitkovich, Brianna and Yu, Tianhe and Xu, Sichun and Xu, Peng and Xiao, Ted and Xia, Fei and Wu, Jialin and Wohlhart, Paul and Welker, Stefan and Wahid, Ayzaan and others},
  booktitle={Conference on Robot Learning},
  pages={2165--2183},
  year={2023},
  organization={PMLR}
}

@article{pertsch2025fast,
  title={Fast: Efficient action tokenization for vision-language-action models},
  author={Pertsch, Karl and Stachowicz, Kyle and Ichter, Brian and Driess, Danny and Nair, Suraj and Vuong, Quan and Mees, Oier and Finn, Chelsea and Levine, Sergey},
  journal={arXiv preprint arXiv:2501.09747},
  year={2025}
}

@article{li2025simplevla,
  title={Simplevla-rl: Scaling vla training via reinforcement learning},
  author={Li, Haozhan and Zuo, Yuxin and Yu, Jiale and Zhang, Yuhao and Yang, Zhaohui and Zhang, Kaiyan and Zhu, Xuekai and Zhang, Yuchen and Chen, Tianxing and Cui, Ganqu and others},
  journal={arXiv preprint arXiv:2509.09674},
  year={2025}
}

@article{zang2025rlinf,
  title={Rlinf-vla: A unified and efficient framework for vla+ rl training},
  author={Zang, Hongzhi and Wei, Mingjie and Xu, Si and Wu, Yongji and Guo, Zhen and Wang, Yuanqing and Lin, Hao and Shi, Liangzhi and Xie, Yuqing and Xu, Zhexuan and others},
  journal={arXiv preprint arXiv:2510.06710},
  year={2025}
}

@article{zang2026rlinf,
  title={RLinf-USER: A Unified and Extensible System for Real-World Online Policy Learning in Embodied AI},
  author={Zang, Hongzhi and Yu, Shu'ang and Lin, Hao and Zhou, Tianxing and Huang, Zefang and Guo, Zhen and Xu, Xin and Zhou, Jiakai and Sheng, Yuze and Zhang, Shizhe and others},
  journal={arXiv preprint arXiv:2602.07837},
  year={2026}
}

@article{shi2026beyond,
  title={Beyond Imitation: Reinforcement Learning-Based Sim-Real Co-Training for VLA Models},
  author={Shi, Liangzhi and Chen, Shuaihang and Gao, Feng and Chen, Yinuo and Chen, Kang and Zhang, Tonghe and Zang, Hongzhi and Zhang, Weinan and Yu, Chao and Wang, Yu},
  journal={arXiv preprint arXiv:2602.12628},
  year={2026}
}

@article{liu2025flow,
  title={Flow-grpo: Training flow matching models via online rl},
  author={Liu, Jie and Liu, Gongye and Liang, Jiajun and Li, Yangguang and Liu, Jiaheng and Wang, Xintao and Wan, Pengfei and Zhang, Di and Ouyang, Wanli},
  journal={arXiv preprint arXiv:2505.05470},
  year={2025}
}

@article{intelligence2026pi,
  title={$\pi_{0.7}$: A Steerable Generalist Robotic Foundation Model with Emergent Capabilities},
  author={Intelligence, Physical and Ai, Bo and Amin, Ali and Aniceto, Raichelle and Balakrishna, Ashwin and Balke, Greg and Black, Kevin and Bokinsky, George and Cao, Shihao and Charbonnier, Thomas and others},
  journal={arXiv preprint arXiv:2604.15483},
  year={2026}
}

@article{wagenmaker2025steering,
  title={Steering your diffusion policy with latent space reinforcement learning},
  author={Wagenmaker, Andrew and Nakamoto, Mitsuhiko and Zhang, Yunchu and Park, Seohong and Yagoub, Waleed and Nagabandi, Anusha and Gupta, Abhishek and Levine, Sergey},
  journal={arXiv preprint arXiv:2506.15799},
  year={2025}
}

@article{lu2025vla,
  title={Vla-rl: Towards masterful and general robotic manipulation with scalable reinforcement learning},
  author={Lu, Guanxing and Guo, Wenkai and Zhang, Chubin and Zhou, Yuheng and Jiang, Haonan and Gao, Zifeng and Tang, Yansong and Wang, Ziwei},
  journal={arXiv preprint arXiv:2505.18719},
  year={2025}
}

@article{luo2025precise,
  title={Precise and dexterous robotic manipulation via human-in-the-loop reinforcement learning},
  author={Luo, Jianlan and Xu, Charles and Wu, Jeffrey and Levine, Sergey},
  journal={Science Robotics},
  volume={10},
  number={105},
  pages={eads5033},
  year={2025},
  publisher={American Association for the Advancement of Science}
}

@article{team2026gigabrain,
  title={Gigabrain-0.5 m*: a vla that learns from world model-based reinforcement learning},
  author={Team, GigaBrain and Wang, Boyuan and Li, Bohan and Ni, Chaojun and Huang, Guan and Zhao, Guosheng and Li, Hao and Li, Jie and Lv, Jindi and Liu, Jingyu and others},
  journal={arXiv preprint arXiv:2602.12099},
  year={2026}
}

@article{littman2001predictive,
  title={Predictive representations of state},
  author={Littman, Michael and Sutton, Richard S},
  journal={Advances in neural information processing systems},
  volume={14},
  year={2001}
}

@article{tan2025interactive,
  title={Interactive post-training for vision-language-action models},
  author={Tan, Shuhan and Dou, Kairan and Zhao, Yue and Kr{\"a}henb{\"u}hl, Philipp},
  journal={arXiv preprint arXiv:2505.17016},
  year={2025}
}

@article{haarnoja2018soft,
  title={Soft actor-critic algorithms and applications},
  author={Haarnoja, Tuomas and Zhou, Aurick and Hartikainen, Kristian and Tucker, George and Ha, Sehoon and Tan, Jie and Kumar, Vikash and Zhu, Henry and Gupta, Abhishek and Abbeel, Pieter and others},
  journal={arXiv preprint arXiv:1812.05905},
  year={2018}
}

@article{kostrikov2021offline,
  title={Offline reinforcement learning with implicit q-learning},
  author={Kostrikov, Ilya and Nair, Ashvin and Levine, Sergey},
  journal={arXiv preprint arXiv:2110.06169},
  year={2021}
}

@inproceedings{jiang2017contextual,
  title={Contextual decision processes with low bellman rank are pac-learnable},
  author={Jiang, Nan and Krishnamurthy, Akshay and Agarwal, Alekh and Langford, John and Schapire, Robert E},
  booktitle={International Conference on Machine Learning},
  pages={1704--1713},
  year={2017},
  organization={PMLR}
}

@inproceedings{liu2022partially,
  title={When is partially observable reinforcement learning not scary?},
  author={Liu, Qinghua and Chung, Alan and Szepesv{\'a}ri, Csaba and Jin, Chi},
  booktitle={Conference on Learning Theory},
  pages={5175--5220},
  year={2022},
  organization={PMLR}
}

@article{subramanian2022approximate,
  title={Approximate information state for approximate planning and reinforcement learning in partially observed systems},
  author={Subramanian, Jayakumar and Sinha, Amit and Seraj, Raihan and Mahajan, Aditya},
  journal={Journal of Machine Learning Research},
  volume={23},
  number={12},
  pages={1--83},
  year={2022}
}

@article{singh2012predictive,
  title={Predictive state representations: A new theory for modeling dynamical systems},
  author={Singh, Satinder and James, Michael and Rudary, Matthew},
  journal={arXiv preprint arXiv:1207.4167},
  year={2012}
}

@inproceedings{singh2003learning,
  title={Learning predictive state representations},
  author={Singh, Satinder P and Littman, Michael L and Jong, Nicholas K and Pardoe, David and Stone, Peter},
  booktitle={Proceedings of the 20th International Conference on Machine Learning (ICML-03)},
  pages={712--719},
  year={2003}
}

@article{laskin2020reinforcement,
  title={Reinforcement learning with augmented data},
  author={Laskin, Misha and Lee, Kimin and Stooke, Adam and Pinto, Lerrel and Abbeel, Pieter and Srinivas, Aravind},
  journal={Advances in neural information processing systems},
  volume={33},
  pages={19884--19895},
  year={2020}
}

@article{shang2021reinforcement,
  title={Reinforcement learning with latent flow},
  author={Shang, Wenling and Wang, Xiaofei and Srinivas, Aravind and Rajeswaran, Aravind and Gao, Yang and Abbeel, Pieter and Laskin, Misha},
  journal={Advances in Neural Information Processing Systems},
  volume={34},
  pages={22171--22183},
  year={2021}
}

@article{mnih2015human,
  title={Human-level control through deep reinforcement learning},
  author={Mnih, Volodymyr and Kavukcuoglu, Koray and Silver, David and Rusu, Andrei A and Veness, Joel and Bellemare, Marc G and Graves, Alex and Riedmiller, Martin and Fidjeland, Andreas K and Ostrovski, Georg and others},
  journal={nature},
  volume={518},
  number={7540},
  pages={529--533},
  year={2015},
  publisher={Nature Publishing Group}
}

@inproceedings{efroni2022provable,
  title={Provable reinforcement learning with a short-term memory},
  author={Efroni, Yonathan and Jin, Chi and Krishnamurthy, Akshay and Miryoosefi, Sobhan},
  booktitle={International Conference on Machine Learning},
  pages={5832--5850},
  year={2022},
  organization={PMLR}
}

@article{schwarzer2020data,
  title={Data-efficient reinforcement learning with self-predictive representations},
  author={Schwarzer, Max and Anand, Ankesh and Goel, Rishab and Hjelm, R Devon and Courville, Aaron and Bachman, Philip},
  journal={International Conference on Learning Representations},
  year={2020}
}

@article{balestriero2025lejepa,
  title={Lejepa: Provable and scalable self-supervised learning without the heuristics},
  author={Balestriero, Randall and LeCun, Yann},
  journal={arXiv preprint arXiv:2511.08544},
  year={2025}
}

@inproceedings{yarats2021improving,
  title={Improving sample efficiency in model-free reinforcement learning from images},
  author={Yarats, Denis and Zhang, Amy and Kostrikov, Ilya and Amos, Brandon and Pineau, Joelle and Fergus, Rob},
  booktitle={Proceedings of the aaai conference on artificial intelligence},
  volume={35},
  number={12},
  pages={10674--10681},
  year={2021}
}

@article{ahuja2022weakly,
  title={Weakly supervised representation learning with sparse perturbations},
  author={Ahuja, Kartik and Hartford, Jason S and Bengio, Yoshua},
  journal={Advances in Neural Information Processing Systems},
  volume={35},
  pages={15516--15528},
  year={2022}
}

@inproceedings{hausknecht2015deep,
  title={Deep Recurrent Q-Learning for Partially Observable MDPs.},
  author={Hausknecht, Matthew J and Stone, Peter},
  booktitle={AAAI fall symposia},
  volume={45},
  pages={141},
  year={2015}
}

@article{chen2021decision,
  title={Decision transformer: Reinforcement learning via sequence modeling},
  author={Chen, Lili and Lu, Kevin and Rajeswaran, Aravind and Lee, Kimin and Grover, Aditya and Laskin, Misha and Abbeel, Pieter and Srinivas, Aravind and Mordatch, Igor},
  journal={Advances in neural information processing systems},
  volume={34},
  pages={15084--15097},
  year={2021}
}

@article{achiam2023gpt,
  title={Gpt-4 technical report},
  author={Achiam, Josh and Adler, Steven and Agarwal, Sandhini and Ahmad, Lama and Akkaya, Ilge and Aleman, Florencia Leoni and Almeida, Diogo and Altenschmidt, Janko and Altman, Sam and Anadkat, Shyamal and others},
  journal={arXiv preprint arXiv:2303.08774},
  year={2023}
}

@article{team2023gemini,
  title={Gemini: a family of highly capable multimodal models},
  author={Team, Gemini and Anil, Rohan and Borgeaud, Sebastian and Alayrac, Jean-Baptiste and Yu, Jiahui and Soricut, Radu and Schalkwyk, Johan and Dai, Andrew M and Hauth, Anja and Millican, Katie and others},
  journal={arXiv preprint arXiv:2312.11805},
  year={2023}
}

@article{driess2023palm,
  title={Palm-e: An embodied multimodal language model},
  author={Driess, Danny and Xia, Fei and Sajjadi, Mehdi SM and Lynch, Corey and Chowdhery, Aakanksha and Ichter, Brian and Wahid, Ayzaan and Tompson, Jonathan and Vuong, Quan and Yu, Tianhe and others},
  journal={arXiv preprint arXiv:2303.03378},
  year={2023}
}

@article{astrom1965optimal,
  title={Optimal control of Markov decision processes with incomplete state estimation},
  author={Astrom, Karl J},
  journal={J. Math. Anal. Applic.},
  volume={10},
  pages={174--205},
  year={1965}
}

@article{smallwood1973optimal,
  title={The optimal control of partially observable Markov processes over a finite horizon},
  author={Smallwood, Richard D and Sondik, Edward J},
  journal={Operations research},
  volume={21},
  number={5},
  pages={1071--1088},
  year={1973},
  publisher={INFORMS}
}

@article{wang2019robust,
  title={Robust reinforcement learning in POMDPs with incomplete and noisy observations},
  author={Wang, Yuhui and He, Hao and Tan, Xiaoyang},
  journal={arXiv preprint arXiv:1902.05795},
  year={2019}
}

@article{galesloot2025robust,
  title={Robust finite-memory policy gradients for hidden-model POMDPs},
  author={Galesloot, Maris FL and Andriushchenko, Roman and {\v{C}}e{\v{s}}ka, Milan and Junges, Sebastian and Jansen, Nils},
  journal={arXiv preprint arXiv:2505.09518},
  year={2025}
}

@article{shi2025memoryvla,
  title={Memoryvla: Perceptual-cognitive memory in vision-language-action models for robotic manipulation},
  author={Shi, Hao and Xie, Bin and Liu, Yingfei and Sun, Lin and Liu, Fengrong and Wang, Tiancai and Zhou, Erjin and Fan, Haoqiang and Zhang, Xiangyu and Huang, Gao},
  journal={arXiv preprint arXiv:2508.19236},
  year={2025}
}

@article{ye2026world,
  title={World action models are zero-shot policies},
  author={Ye, Seonghyeon and Ge, Yunhao and Zheng, Kaiyuan and Gao, Shenyuan and Yu, Sihyun and Kurian, George and Indupuru, Suneel and Tan, You Liang and Zhu, Chuning and Xiang, Jiannan and others},
  journal={arXiv preprint arXiv:2602.15922},
  year={2026}
}

@article{kim2026cosmos,
  title={Cosmos policy: Fine-tuning video models for visuomotor control and planning},
  author={Kim, Moo Jin and Gao, Yihuai and Lin, Tsung-Yi and Lin, Yen-Chen and Ge, Yunhao and Lam, Grace and Liang, Percy and Song, Shuran and Liu, Ming-Yu and Finn, Chelsea and others},
  journal={arXiv preprint arXiv:2601.16163},
  year={2026}
}

@article{koo2025hamlet,
  title={Hamlet: Switch your vision-language-action model into a history-aware policy},
  author={Koo, Myungkyu and Choi, Daewon and Kim, Taeyoung and Lee, Kyungmin and Kim, Changyeon and Seo, Younggyo and Shin, Jinwoo},
  journal={arXiv preprint arXiv:2510.00695},
  year={2025}
}

@article{li2025cronusvla,
  title={Cronusvla: Transferring latent motion across time for multi-frame prediction in manipulation},
  author={Li, Hao and Yang, Shuai and Chen, Yilun and Tian, Yang and Yang, Xiaoda and Chen, Xinyi and Wang, Hanqing and Wang, Tai and Zhao, Feng and Lin, Dahua and others},
  journal={arXiv e-prints},
  pages={arXiv--2506},
  year={2025}
}

@article{bi2025motus,
  title={Motus: A unified latent action world model},
  author={Bi, Hongzhe and Tan, Hengkai and Xie, Shenghao and Wang, Zeyuan and Huang, Shuhe and Liu, Haitian and Zhao, Ruowen and Feng, Yao and Xiang, Chendong and Rong, Yinze and others},
  journal={arXiv preprint arXiv:2512.13030},
  year={2025}
}

@article{li2026causal,
  title={Causal World Modeling for Robot Control},
  author={Li, Lin and Zhang, Qihang and Luo, Yiming and Yang, Shuai and Wang, Ruilin and Han, Fei and Yu, Mingrui and Gao, Zelin and Xue, Nan and Zhu, Xing and others},
  journal={arXiv preprint arXiv:2601.21998},
  year={2026}
}

@article{yuan2026fast,
  title={Fast-WAM: Do World Action Models Need Test-time Future Imagination?},
  author={Yuan, Tianyuan and Dong, Zibin and Liu, Yicheng and Zhao, Hang},
  journal={arXiv preprint arXiv:2603.16666},
  year={2026}
}

@article{maes2026leworldmodel,
  title={Leworldmodel: Stable end-to-end joint-embedding predictive architecture from pixels},
  author={Maes, Lucas and Lidec, Quentin Le and Scieur, Damien and LeCun, Yann and Balestriero, Randall},
  journal={arXiv preprint arXiv:2603.19312},
  year={2026}
}

@inproceedings{radford2021learning,
  title={Learning transferable visual models from natural language supervision},
  author={Radford, Alec and Kim, Jong Wook and Hallacy, Chris and Ramesh, Aditya and Goh, Gabriel and Agarwal, Sandhini and Sastry, Girish and Askell, Amanda and Mishkin, Pamela and Clark, Jack and others},
  booktitle={International conference on machine learning},
  pages={8748--8763},
  year={2021},
  organization={PmLR}
}

@inproceedings{perez2018film,
  title={Film: Visual reasoning with a general conditioning layer},
  author={Perez, Ethan and Strub, Florian and De Vries, Harm and Dumoulin, Vincent and Courville, Aaron},
  booktitle={Proceedings of the AAAI conference on artificial intelligence},
  volume={32},
  number={1},
  year={2018}
}

\newpage
\appendix
\section{How does the weight allocation of training objectives affect performance?}
\label{exp:lambda-abl}
\begin{wrapfigure}{r}{0.5\textwidth}
\vspace{-10pt}
\centering
\includegraphics[width=\linewidth]{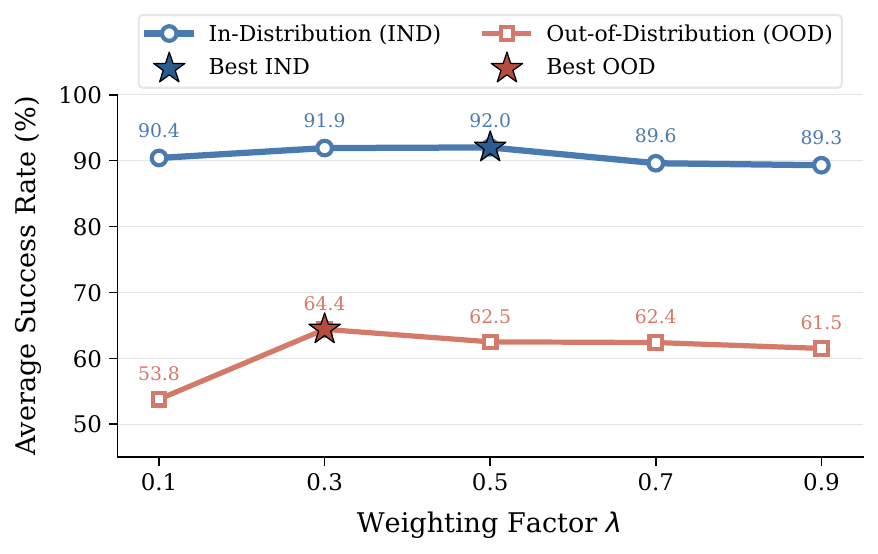}
\vspace{-20pt}
\caption{Effect of $\lambda$ on IND and OOD performance. The plot shows the average success rate for both IND (blue) and OOD (red) across different values of $\lambda$, with stars indicating the best performance for each metric.}
\label{fig:lambda_ablation}
\vspace{-30pt}
\end{wrapfigure}

Another interesting question concerns the hyperparameter $\lambda$ in our method, which balances the original value regression objective and the world prediction objective. Specifically, $\lambda = 0$ uses only the value objective, while $\lambda = 1$ assigns equal weight to both. Results are shown in Figure~\ref{fig:lambda_ablation}. When the supervision signal is weak, even if IND performance remains reasonable, OOD performance is comparable to the original $\pi_{\text{RL}}$ baseline. In contrast, with strong supervision, OOD performance remains competitive even if IND is slightly compromised.

(1) $\lambda$ inherently controls a trade-off between world prediction and value learning. Thus, neither extremely large nor small $\lambda$ yields optimal results; there exists a stable intermediate range, with the best IND and OOD performance both occurring within $[0.3, 0.5]$.

(2) $\lambda = 0.9$ (dominated by the world prediction objective) achieves better OOD performance than $\lambda = 0.1$ (dominated by the value regression objective), indicating that the world prediction objective contributes positively to generalization.

(3) OOD performance is more sensitive to $\lambda$. Varying $\lambda$ leads to a fluctuation of $10.6$ percentage points in OOD, compared to only $2.7$ percentage points in IND.

\section{Can the trained policy achieve sim-to-real transfer?}
\label{exp:sim2real}
Since RL methods optimize for reward maximization, they are generally expected to achieve better generalization than SFT. To investigate this, we conduct two experiments, as shown in Table~\ref{tab:sim2real}:

(1) We take (i) the SFT checkpoint trained on 16,800 simulated samples and (ii) the RL checkpoint obtained from 16,800 SFT samples followed by 285 RL steps (using an additional 18,240 simulated samples as batch size of a step is set to 64). Both models are deployed directly in the real world on a pick-and-place task, using similar camera view as in the simulation, and we compare their success rates.
Among them, the SFT data in the simulation is generated by a rule-based planner, resulting in a form of data that supports parallel processing, low cost, and large-scale generation. In contrast, the rollout data is relatively more expensive to obtain.

We surprisingly find that the SFT model is completely unable to successfully grasp any object in the real world, despite our attempts with various different placement positions. In contrast, the policy obtained after RL achieves a high probability of successful grasping and placing in the central region of the camera's field of view, while exhibiting significant grasping deviations at the edges of the field of view.

(2) We start from a model fine-tuned with real-world SFT data (carrot pick-and-place) and continue RL training in simulation, examining whether this leads to improved success rates.

\begin{table}[t]
\centering
\small
\caption{
Performance comparison under Sim IND, Sim OOD, and Real settings. The Sim OOD results are further broken down into vision, semantic, and execution dimensions. All methods are trained via RL on top of SFT initialization, and we report the best performance after 1,000 training steps. Bold denotes the best performance.
}
\setlength{\tabcolsep}{5pt}
\renewcommand{\arraystretch}{1.2}
\begin{tabular}{l l c|c c c c|c c c}
\toprule
\multirow{2}{*}{Model} & \multirow{2}{*}{Method}
& \multicolumn{1}{c}{Sim-IND} 
& \multicolumn{4}{c}{Sim-OOD} 
& \multicolumn{3}{c}{Real-Pick up object} \\
\cmidrule(lr){3-3} \cmidrule(lr){4-7} \cmidrule(lr){8-10}
& & avg. & vision & semantic & execution & avg. & carrot & banana & pepper \\
\midrule
\multirow{4}{*}{$\pi_{0.5}$}
& sim SFT & 47.0 & 40.2 & 16.6 & 22.4 & 26.4 & 0/25 & 0/25 & 0/25 \\
& + sim RL & 91.9 & 78.1 & 58.5 & 56.5 & 64.4 & 7/25 & 7/25 & 6/25 \\
& Real SFT & 6.9 & 5.1 & 5.9 & 5.2 & 5.4 & 13/25 & 2/25 & 4/25 \\
& + sim RL & 73.5 & 42.1 & 35.9 & 35.8 & 37.9 & 11/25 & 8/25 & 9/25 \\
\bottomrule
\end{tabular}
\label{tab:sim2real}
\vspace{-12pt}
\end{table}

As the results shown, after fine-tuning with 50 real-world data samples, the policy achieves a success rate of only 6.9\% in simulation. This result stands in stark contrast to the sim-to-real findings, where even training with 16,800 simulation samples fails to produce a policy capable of successfully performing the task in the real world.

We believe the above results are likely consistent with existing perspectives in the field regarding simulation data versus real-robot data: although simulation data is cheap and can be generated in large quantities, the simulation environment is completely idealized. Even if we introduce rule-based perturbations (such as changing backgrounds, objects, and positions, as done in our experiments), it remains difficult to obtain noise or disturbances similar to those in real environments (e.g., motor temperature variations during inference), thereby making it difficult to provide corresponding robustness. Therefore, directly using simulation data for SFT may not yield significant OOD performance gains. This is reflected in our experiments, where simulation SFT failed to complete even a single pick-and-place task on the real robot, not even grasping succeeded. In contrast, using only 50 real-world data points enabled task completion in simulation. Moreover, since RL optimizes for reward maximization, it may offer better OOD performance compared to directly learning expert actions.

Furthermore, although we train the policy exclusively on the task of grasping carrot and placing it on plate, it still has a non-zero probability of successfully grasping bananas and peppers. After 1000 steps of reinforcement learning in simulation, the model's performance on grasping carrots slightly degrades, while its performance on grasping other objects significantly improves.

This is a interesting phenomenon. One possible explanation is that, in order to adapt to the simulation environment, the model adjusts its feature representation. This adjustment process may make the model more sensitive to features that are more common or easier to grasp in simulation, and these features happen to be more compatible with general-purpose grasping rather than with the in-distribution carrot in the simulation environment. This may also cause the model to partially forget its original ability to grasp carrots.

\section{Does Critic Model Affect Generalization?}
\label{exp:critic}
We investigate the impact of the critic model on generalization performance. $\pi$-StepNFT highlights that PPO’s scalar return-based critic is prone to overfitting. Based on this, we explore 2 questions:

\begin{wrapfigure}{r}{0.45\textwidth}
\centering
\includegraphics[width=0.8\linewidth]{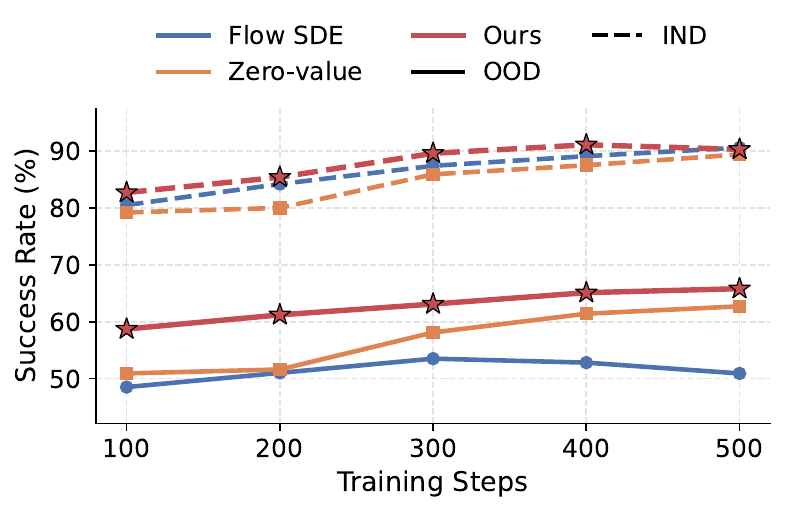}
\caption{Training dynamics of Flow-SDE (dropping phenomenon) and Zero-Value.}
\label{fig:zero_value}
\vspace{-15pt}
\end{wrapfigure}

\textbf{(1)} Whether OOD performance improves initially but deteriorates as the critic overfits.
\textbf{(2)} What if we remove the critic in Flow-SDE by fixing the value function to $V(o) = 0$ for all observations, will this improves OOD performance?

We present the results in Figure~\ref{fig:zero_value}. Key findings:
\textbf{(1)} OOD performance of Flow-SDE shows a ``dropping phenomenon'', indicating overfitting in later stages.  
\textbf{(2)} Zero-value ablation achieves better OOD performance than Flow-SDE at comparable IND levels, suggesting overfitting may be linked to the critic model.
\textbf{(3)} No overfitting is observed in WCM during the first 500 steps, though we do not claim it is entirely immune.

\section{Details for On-policy and Off-policy Training}
\label{app:appendix_algorithms}

\subsection{Algorithms}
We present the detailed algorithms for on-policy and off-policy pipelines. The foundational algorithms we selected all require a critic model or a value function. Specifically, for on-policy learning, we use PPO for the autoregressive model and Flow-SDE (an enhancement of PPO tailored for flow-matching models) for the flow-matching model. SIGReg is not adopted in the on-policy pipeline, as constraining the VLM latent would introduce unnecessary computational overhead. For off-policy, we use AWR for the AR model and RECAP for the flow-matching model. We leverage WCM to replace the origin critic model or value function in the above algorithms, thus improving value estimation by adding historical observation. Below are the procedures for the training pipelines, where each method updates both the policy and WCM parameters through gradient-based optimization.

\begin{algorithm}[ht]
\caption{PPO with WCM Update}
\label{algo:ppo_wcm_update}
\KwIn{
  Initial policy $\pi_{\theta}$ , initial WCM $W_\phi$, language instruction $\ell$. \\
    Hyperparameters: loss weight $\lambda$ , learning rate $\alpha$, number of steps $U$ , batch size $B$ , observation history length $K$. \\
}
\KwOut{
  Updated policy $ \pi_{\theta'} $, updated WCM $ V_{\phi'} $.
}
\For{each optimization step $ u = 1, 2, \dots, U $}{
    \For{each episode}{
        Collect trajectory data $ \{o_t, a_t, r_t, z_t, \hat{z}_{t+1}, \hat{V}_{t}, \dots \} $ using $ \pi_\theta $ and $ W_\phi $ (Equa.(\ref{qua1})-(\ref{qua3}))
    }
    \For{each batch of size $ B $}{
        Compute the policy gradient loss:
        \begin{equation}
        \mathcal{L}_{\text{actor}} = - \mathbb{E}_t \left[ \min \left( \rho_t(\pi) \hat{A}_t, \text{clip}(\rho_t(\pi), 1-\epsilon, 1+\epsilon) \hat{A}_t \right) \right]
        \end{equation} \\
        where $ \rho_t(\pi) = \frac{\pi(a_t | o_t)}{\pi_{\text{old}}(a_t | o_t)} $  and $ \hat{A}_t $ is the estimated advantage using GAE.
        Compute the total critic loss from Equa.(\ref{eq:4}) and (\ref{eq:7}):
        \begin{equation}
        \mathcal{L} = \mathcal{L}_{\text{value}} + \lambda \cdot \mathcal{L}_{\text{pred}}
        \end{equation} \\
        Update policy and WCM via stochastic gradient descent:
        \begin{equation}
        \phi' = \phi - \alpha \nabla_\phi \mathcal{L}
        \end{equation}
        \begin{equation}
        \theta' = \theta - \alpha \nabla_\theta \mathcal{L}_{\text{actor}}
        \end{equation}\\
    }
}
\end{algorithm}

\begin{algorithm}[ht]
\caption{AWR with WCM Update}
\label{algo:awr}
\KwIn{
  Initial policy $ \pi_{\theta} $, WCM $ W_\phi $, Trajectory Buffer $ \mathcal{B}  = \varnothing $, language instruction $ \ell $. \\
    Hyperparameters: loss weight $ \lambda $, $ \eta $, learning rate $ \alpha $, number of iterations $ I $, batch size $ B $, observation history length $ K $
}
\KwOut{
  Updated policy $ \pi_{\theta'} $, updated WCM $ V_{\phi'} $.
}
\For{each episode}{
    Collect trajectory $ \{o_t, a_t, r_t, z_t, \hat{z}_{t+1}, G_t, \dots \} $ by teleop and add to $ \mathcal{B} $ (Equa.(\ref{qua1})-(\ref{qua3}) (\ref{qua5}) )
}
\For{each iteration $ i = 1, 2, \dots, I $}{
    \For{each batch of size $ B $}{
        Compute critic loss from Equa.(\ref{eq:4})-(\ref{eq:9}) and update WCM via stochastic gradient descent:
        \begin{equation}
        \phi' = \phi - \alpha \nabla_\phi \mathcal{L}
        \end{equation}\\
    }
    \For{each batch of size $ B $}{
        Compute the AWR loss for the policy update:
        \begin{equation}
        \mathcal{L}_{\text{actor}} = \mathbb{E}_t \left[ - \log \pi(a_t | o_t) \exp\left(\frac{1}{\beta} \left( G_t - V(o_{t-K+1:t}) \right)\right) \right]
        \end{equation}\\
        Update policy via stochastic gradient descent:
        \begin{equation}
        \theta' = \theta - \alpha \nabla_\theta \mathcal{L}_{\text{actor}}
        \end{equation}\\
    }
    Collect trajectory data $ \{o_t, a_t, r_t, z_t, \hat{z}_{t+1}, G_t, \dots \} $ by policy rollout and add to $ \mathcal{B} $.
}
\end{algorithm}

\begin{algorithm}[ht]
\caption{RECAP with WCM Update}
\label{algo:recap}
\KwIn{
  Initial policy $ \pi_{\theta} $, WCM $ W_\phi $, Trajectory Buffer $ \mathcal{B}  = \varnothing $, language instruction $ \ell $. \\
    Hyperparameters: loss weight $ \lambda $, $ \eta $, $ \beta $, learning rate $ \alpha $, number of iterations $ I $, batch size $ B $, observation history length $ K $, discount factor $ \gamma $, advantage threshold $ \epsilon_\ell $, TD target step $ N $.
}
\KwOut{
  Updated policy $ \pi_{\theta'} $, updated WCM $ V_{\phi'} $.
}
\For{each episode}{
    Collect trajectory $ \{o_t, a_t, r_t, z_t, \hat{z}_{t+1}, G_t, \dots \} $ by teleop and add to $ \mathcal{B} $ (Equa.(\ref{qua1})-(\ref{qua3}) (\ref{qua5}) )
}
\For{each iteration $ i = 1, 2, \dots, I $}{
    \For{each batch of size $ B $}{
        Compute critic loss from Equa.(\ref{eq:4})-(\ref{eq:9}) and update WCM via stochastic gradient descent: 
        \begin{equation} 
        \phi' = \phi - \alpha \nabla_\phi \mathcal{L} 
        \end{equation} \\
    }
    \For{each batch of size $ B $}{
        Estimate value and compute advantage: 
        \begin{equation}
        A(o_t, a_t, \ell) = \text{normalize}(G_{t:t+N}) + \gamma^N \hat{V}(o_{t+N-K+1:t+N}) - \hat{V}(o_{t-K+1:t}).
        \end{equation} \\
        Compute the RECAP loss for the policy update:
        \begin{equation}
        \mathcal{L}_{\text{actor}} = -\log \pi_{\theta}(a_t \mid o_t, \ell) - \beta \log \pi_{\theta}(a_t \mid \mathbf{1}(A(o_t, a_t, \ell) > \epsilon_\ell), o_t, \ell),
        \end{equation}\\
        Update policy via stochastic gradient descent:
        \begin{equation}
        \theta' = \theta - \alpha \nabla_\theta \mathcal{L}_{\text{actor}}
        \end{equation}\\
    }
    Collect trajectory data $ \{o_t, a_t, r_t, z_t, \hat{z}_{t+1}, G_t, \dots \} $ by policy rollout and add to $ \mathcal{B} $.
}
\end{algorithm}

\subsection{Training Details}
\textbf{On-policy training in simulation.} 
Since the simulation environment enables low-cost rollout of large amounts of data, we adopt an on-policy method for training. For the $\pi$~\cite{intelligence2025pi,black2024pi_0} series models, we perform policy updates using Flow-SDE~\cite{chen2025pirl} improved by WCM. Flow-SDE is a method proposed by the RLinf team, which combines DPPO~\cite{ren2024diffusion} and Flow-GRPO~\cite{liu2025flow} to specifically adapt and enhance VLA models, offering strong engineering practicality and reproducibility. In the RLinf implementation, the critic model is a 3-layer MLP; they found that such a lightweight critic can fit task-specific requirements well. We replace it with our WCM, which takes the same input, i.e., the latent representation from the VLM backbone, and outputs a scalar value. For OpenVLA-OFT~\cite{kim2025fine}, we use the PPO implemented by RLinf~\cite{zang2025rlinf} and similarly replace the MLP-based critic with WCM.

On the ManiSkill~\cite{mu2021maniskill} benchmark, to ensure a fair comparison with baseline methods, we adopt exactly the same settings as RL4VLA~\cite{liu2025can}, with 25 pick-and-place tasks and 2 major categories: in-distribution (IND) and out-of-distribution (OOD). IND includes various pick-and-place tasks with diverse objects, while OOD evaluates model performance under perturbations across three dimensions. We use the hyperparameters officially provided by RLinf without any additional tuning to ensure a perfectly fair comparison. Our batch size is set to 64, meaning that at each update step, the model observes 64 complete trajectories. All reported results are obtained after training for a full 1,000 steps, which means the model sees 64,000 trajectories during the RL phase, which is a substantial amount for VLA-RL. Thus, our results essentially reflect the best performance of the model under this setting, as 1,000 steps are sufficient for the model to converge well. The checkpoint used to initialize training is an SFT model trained on 16,800 trajectories generated by a rule-based planner. For OpenVLA-OFT, we also train from a checkpoint that is SFT only on LIBERO-Goal, Object, and Spatial, which achieves only 0.78\% success rate. After about 600 steps of training, we are able to bring its performance close to near-perfect, with strong training stability.

\begin{wrapfigure}{r}{0.45\textwidth}
\vspace{-15pt}
\centering
\includegraphics[width=\linewidth]{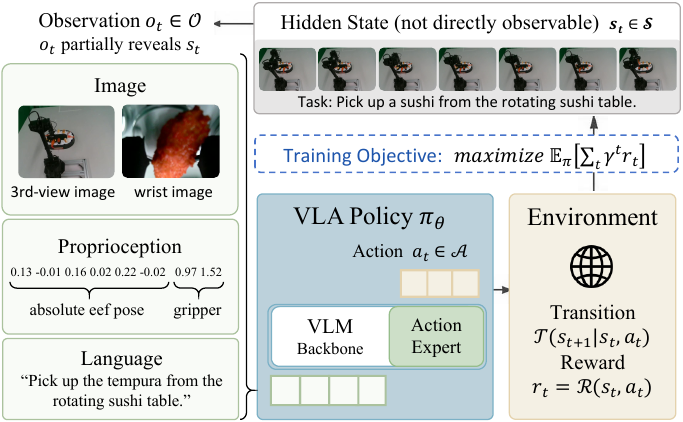}
\caption{Illustration of the rollout process. Starting from an observation $o_t$ partially revealing hidden state $s_t$, the policy $\pi_{\theta}$ generate action $a_t$ and the environment provides the transition and reward.}
\label{fig:new}
\vspace{-15pt}
\end{wrapfigure}

We further evaluate on the MetaWorld~\cite{yu2020meta} and CALVIN~\cite{mees2022calvin} benchmarks. While ManiSkill is limited to pick-and-place tasks, MetaWorld offers a broader variety of tasks, and CALVIN allows us to assess the model's long-horizon capabilities. Again, we follow the same hyperparameters used by RLinf for training and comparison. For MetaWorld, we report the final success rate; for CALVIN, we report the average task completion length, which is the official metric recommended by CALVIN. This metric reflects how far the model can proceed before failing, with a minimum of 0 (unable to complete even one task) and a maximum of 5 (successfully completing all five tasks randomly selected by the benchmark). All simulation evaluations are averaged over three runs, reporting the error bars.

\begin{wrapfigure}{r}{0.45\textwidth}
    \centering
    \includegraphics[width=0.9\linewidth]{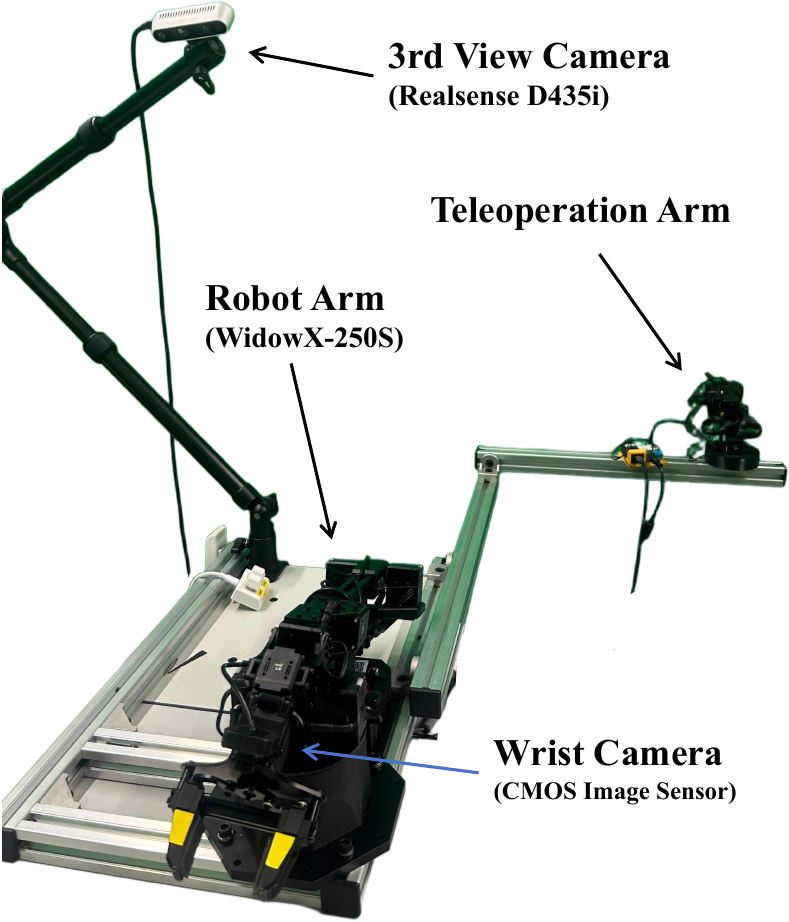}
    \caption{Demonstration of the Real-Robot Experiment Environment.}
    \label{fig:real_robot_demo}
\end{wrapfigure}

\textbf{Off-policy training in real-world.} 
We have also conducted extensive work on real-robot RL to demonstrate the effectiveness of our method in physical environments. Since real-world data is difficult to acquire and rollouts risk damaging the hardware or motors of the manipulator (We set hardware limits to prevent such incidents), we believe that on-policy training is not a wise choice. Instead, we adopt two off-policy RL algorithms that have gained significant traction in the community: AWR~\cite{peng2019advantage} and RECAP~\cite{intelligence2025pi}. The former is a classic offline RL method; however, in our implementation, we do not use an additional transition model, but simply supplement the data buffer with online rollout data, and thus we prefer to call it off-policy rather than offline. The latter is widely known in the community for its use in $\pi^{*}_{0.6}$~\cite{intelligence2025pi}, where it achieved substantial performance improvements. We use AWR to update the OpenVLA-OFT model and RECAP to update the $\pi_{0.5}$ model. It is worth noting that in our implementation, the pipelines of the two methods are largely identical, with the only difference being the loss function used for policy update.

All our real-robot experiments are conducted on the WidowX-250S robot arm, using both third- and wrist-camera views, as shown in Figure~\ref{fig:real_robot_demo}. We employ delta end-effector pose as the control method and use absolute end-effector pose along with a two-finger gripper state (representing openness) as proprioceptive input. We built a master-slave teleoperation data collection pipeline based on ROS2 and collected an initial set of 100 trajectories for each task. We trained our WCM on seven tasks separately, providing a total of approximately 250k transition pairs. For the policy, three pick-and-place tasks share a single SFT policy, two deformable object manipulation tasks share another SFT policy, and the remaining two tasks are each SFT separately. In each round of RL update, we collect all successful rollouts. For unsuccessful rollouts, we apply human-in-the-loop corrections to turn them into successful trajectories (for reward, we assign a large negative reward at the failure point and a reward of 0 at the success point). In each RL update round, 50 trajectories (including both failed and successful ones) are collected per task, and we use new data from each task to update the WCM. $C_{\mathrm{fail}}$ is set to $300$. Each reported data point is obtained after 8 iterations of RL updates.

In our implementations of AWR and RECAP, we selected a value model comprising a SigLip 400M Encoder and a Gemma 270M Backbone guided by the reference $\pi^{*}_{0.6}$~\cite{intelligence2025pi}, while adopting a training paradigm identical to that of our World Critic Model (+WCM). As demonstrated in Table~\ref{tab:real_robot_results}, WCM achieves superior performance over the standard baseline. From a theoretical perspective, this gain is attributed to two factors: first, unlike conventional scalar regression, our WCM incorporates world prediction objectives, fostering an enhanced state representation that more accurately reconstructs the state space; second, by integrating historical observations, our critic model effectively captures temporal dynamics, which is crucial as the Vision-Language-Action (VLA) model continuously moves during manipulation tasks. Empirically, these advantages manifest in distinct behavioral insights across two challenging setups. In deformable object manipulation, single-frame critic models trained via reinforcement learning tend to drive the end-effector into the tabletop, causing motor stalls; this occurs because reaching the tabletop's $z$-coordinate correlates with successful grasping, yet lacking historical context, the single-frame model cannot discern whether this physical obstruction is beneficial or detrimental from a static snapshot. Conversely, our WCM-trained model executes remarkably smooth trajectories with negligible collisions, as reflected in the value curves where frames involving tabletop collisions exhibit a noticeable drop in value, penalizing the policy from learning such actions. Furthermore, in the Conveyor Belt Sushi Picking task, the single-frame critic baseline struggles to improve due to a pronounced post-grasp latency of several seconds that leads to collisions with adjacent objects or items being dragged away; in contrast, the WCM significantly boosts both the success rate and the operational fluidity of the execution sequence.

\begin{wrapfigure}{r}{0.45\textwidth}
    \centering
    \vspace{-10pt}
    \includegraphics[width=0.9\linewidth]{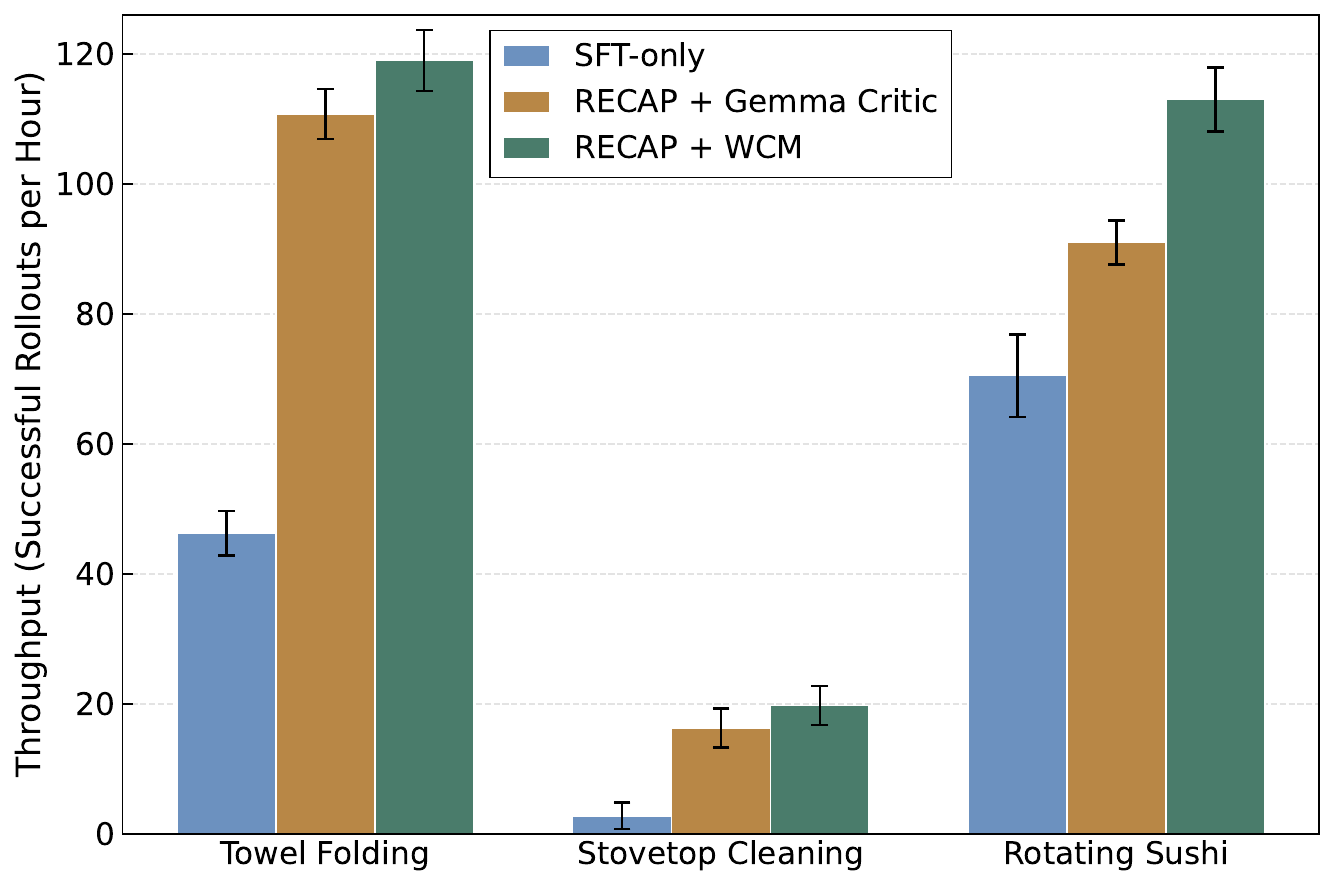}
    \caption{Throughput comparison of different policy variants across three tasks. Throughput is measured as the number of successful rollouts per hour. Bars represent mean values over four evaluation runs, with error bars indicating standard deviation. The SFT-only model shows low throughput across all tasks, while RECAP significantly improves throughput. Among the two RECAP variants, the WCM-based critic consistently outperforms the original Gemma-based critic.}
    \label{fig:real_world_throughput}
    \vspace{-40pt}
\end{wrapfigure}

\textbf{Computational resources and settings.} For simulation, all experimental data points were obtained on an 8 $\times$ H100 machine, including training and evaluation. For real-robot experiments, all training was conducted on 8 $\times$ H100 GPUs, while inference was performed on a local workstation with RTX5090. The control frequency for all real-robot experiments was 10 Hz, and the action chunk size was set to 5. Thus, each action chunk spans 0.5 seconds, with a new observation being received at each control step (every 0.5 s).

\subsection{Inference Throughput}

We evaluated the inference throughput of the $\pi_{0.5}$ model across three tasks: towel folding, stovetop cleaning, and rotating sushi picking. Throughput is measured as the number of successful rollouts per hour. The results are presented in Figure~\ref{fig:real_world_throughput}.

The SFT-only model yields the lowest throughput, due to its low initial success rate and inefficient action execution, characterized by frequent pauses and small-magnitude movements, which prolong each rollout. After RL fine-tuning, throughput improves significantly as the policy becomes both more successful and smoother.

Among the two RECAP-trained variants, the model using the WCM-based critic consistently outperforms the Gemma VLM-based critic variant across all tasks, indicating that the choice of critic model has a substantial impact on inference efficiency.

\subsection{Traning Curve}

\begin{figure}[htbp]
    \centering
    \includegraphics[width=0.85\textwidth]{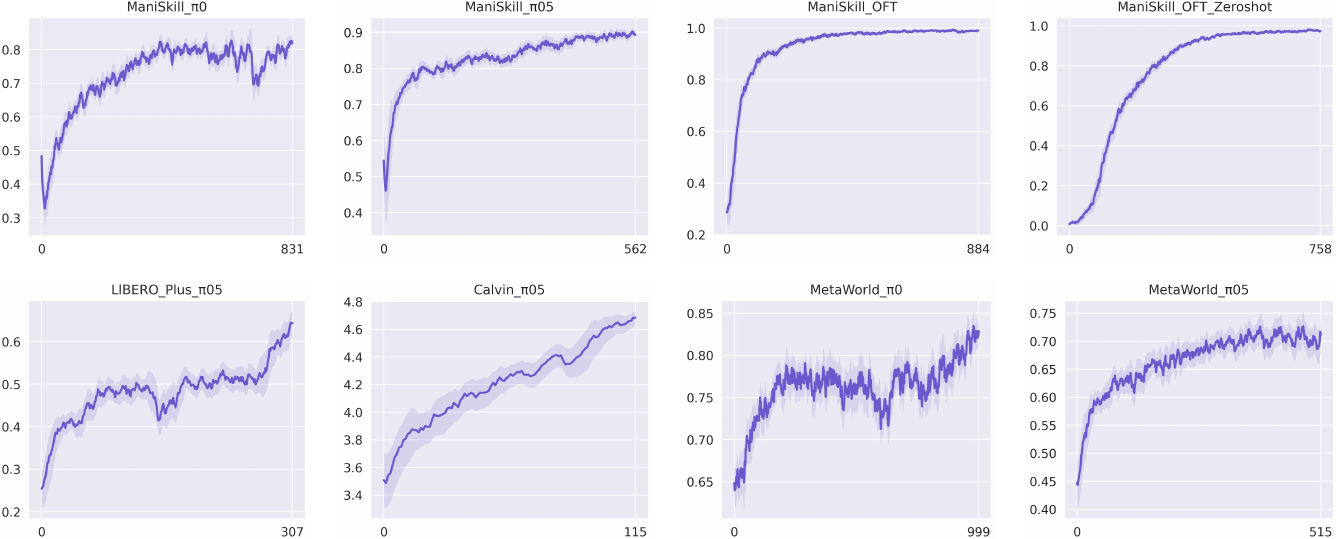}
    \caption{Training curves across 8 different settings, plotted from the initial step (step 0) to the point where the optimal result (as reported in our work) is achieved. It is worth noting that for Maniskill-OOD and LIBERO-Plus, since the test settings differ from the training settings, there will be some numerical discrepancies.}
    \label{fig:training_curve}
\end{figure}

To better understand the optimization behavior and convergence of different configurations, we plot the training curves of all eight settings in Figure \ref{fig:training_curve}. The curves are recorded from the very beginning (step 0) until each setting reaches its reported optimal performance. As shown in the figure, most settings exhibit stable improvement over training steps, though their convergence speeds and final performance vary across configurations. Notably, for Maniskill-OOD and LIBERO-Plus, because the test settings differ from the training settings, the reported numerical results show certain discrepancies compared to the training curves; these differences are expected and reflect the generalization gap under distribution shift.

\section{Value Curve Visualization of WCM}
\label{app:appendix_value_curve}

To better illustrate WCM's contribution to the RL process, we visualize the value curves of successful and failed trajectories in both simulation and real-world tasks. 

\begin{figure}[htbp]
    \centering
    \includegraphics[width=1.0\textwidth]{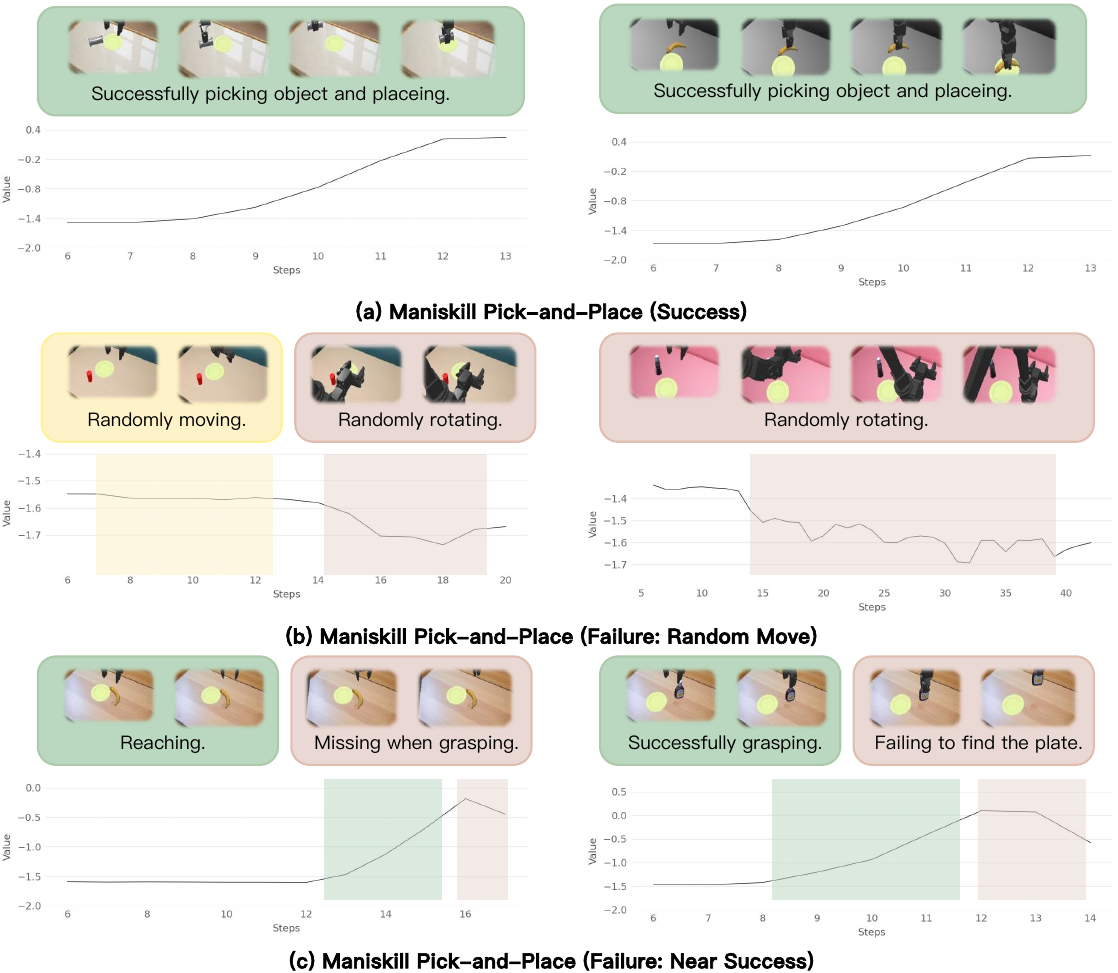}
    \caption{WCM value curve in Maniskill.WCM exhibits strong discriminative capability between successful and failed trajectories. Because the motion planner in the simulation environment supplies ideal successful trajectories, the value estimates for successful trajectories are exceptionally smooth.}
    \label{fig:value_curves_maniskill}
\end{figure}

For simulation, we construct 5k successful trajectories and 2k failed ones. Among the failed trajectories, 1k are random end-effector (EEF) movements (translation + rotation), and the remaining 1k are ``near-success'' trajectories where the planner is perturbed (e.g., adding a coordinate offset to the target grasping position), causing task failure. We train a single WCM on all 7k trajectories and evaluate it on unseen trajectories. The results are shown in Figure~\ref{fig:value_curves_maniskill}. For successful trajectories, the value curve increases monotonically, which is expected given that the trajectories in simulation are idealized and near-optimal. The two failure types exhibit distinct patterns. For random EEF movement, which shows no tendency toward task completion, the value depends primarily on the EEF's position during random wandering; translation causes only slight value degradation as the EEF may occasionally move closer to or farther from the target, while rotation leads to a sharp value drop, as it deviates completely from successful behavior. For near-success trajectories, the value first rises and then falls, mirroring the trajectory's initial progression toward success followed by eventual failure.

\begin{figure}[htbp]
    \centering
    \includegraphics[width=1.0\textwidth]{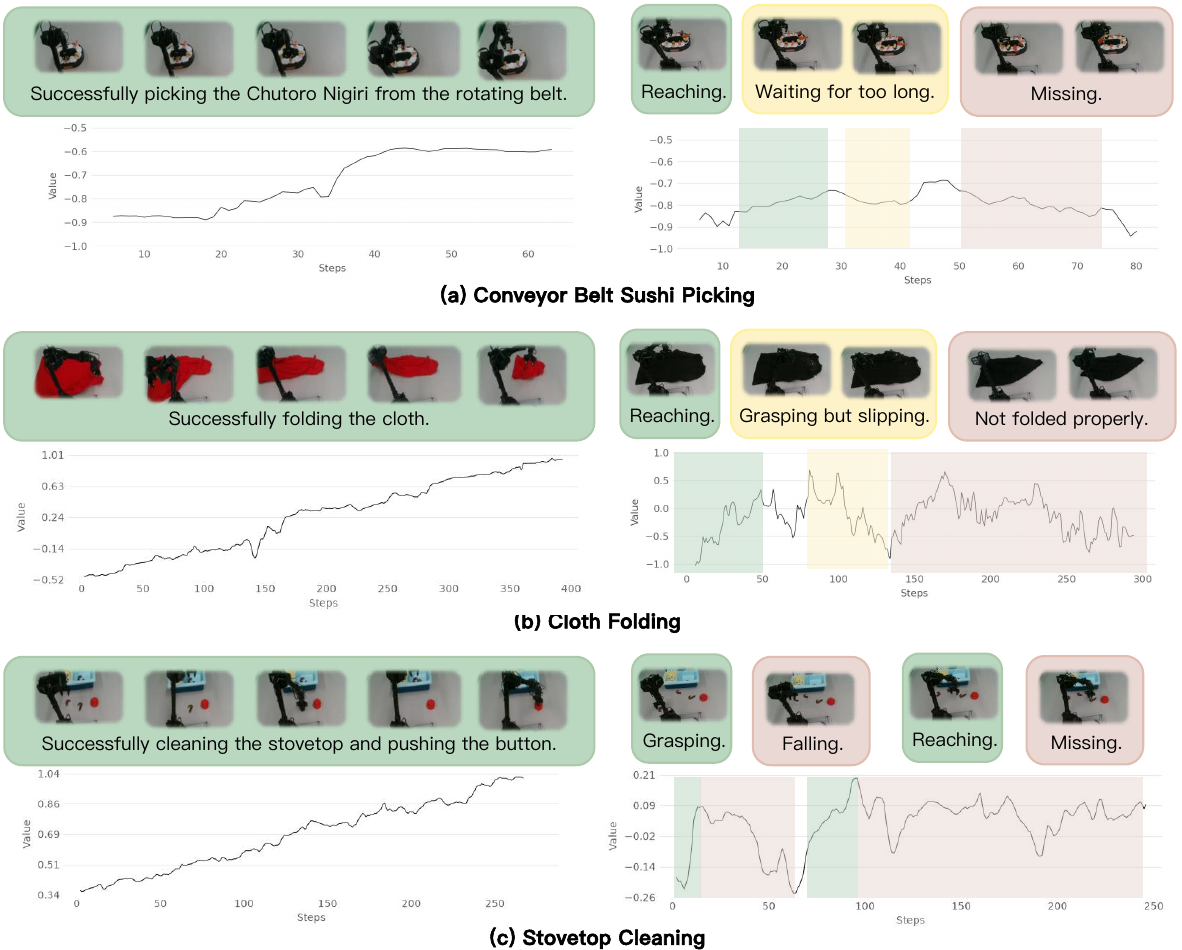}
    \caption{WCM value curve in the real world. Unlike simulation, real-world trajectories are not ideal: teleoperated trajectories are subject to various confounding factors, and both the environment and the camera introduce visual noise. Consequently, even successful trajectories show certain fluctuations. Nevertheless, the discriminability between successful and failed trajectories remains high.}
    \label{fig:value_curves_realworld}
\end{figure}

For real-world tasks, we train WCM separately per task, each with 500 trajectories (including both successes and failures). The results are shown in Figure~\ref{fig:value_curves_realworld}. Due to the less idealized nature of the real-world environment and the inevitable inclusion of suboptimal actions in the collected trajectories, even successful trajectories do not exhibit strictly monotonic value curves. Value stagnation or decline is primarily caused by factors such as pauses during grasping, collisions with the table or objects, unfavorable grasp poses, or imperfect task completion (e.g., misplaced items, misaligned cloth folding). For failed trajectories, the value curves generally exhibit a downward trend.

\section{Additional Generalization Analysis of the Critic Model}
\label{app:appendix_generalization}

We present additional analysis of the generalization capability of the critic model, to answer whether WCM trained on IND data can produce more reasonable value distributions when directly applied to OOD settings.

\begin{figure}[htbp]
    \centering
    \includegraphics[width=0.85\textwidth]{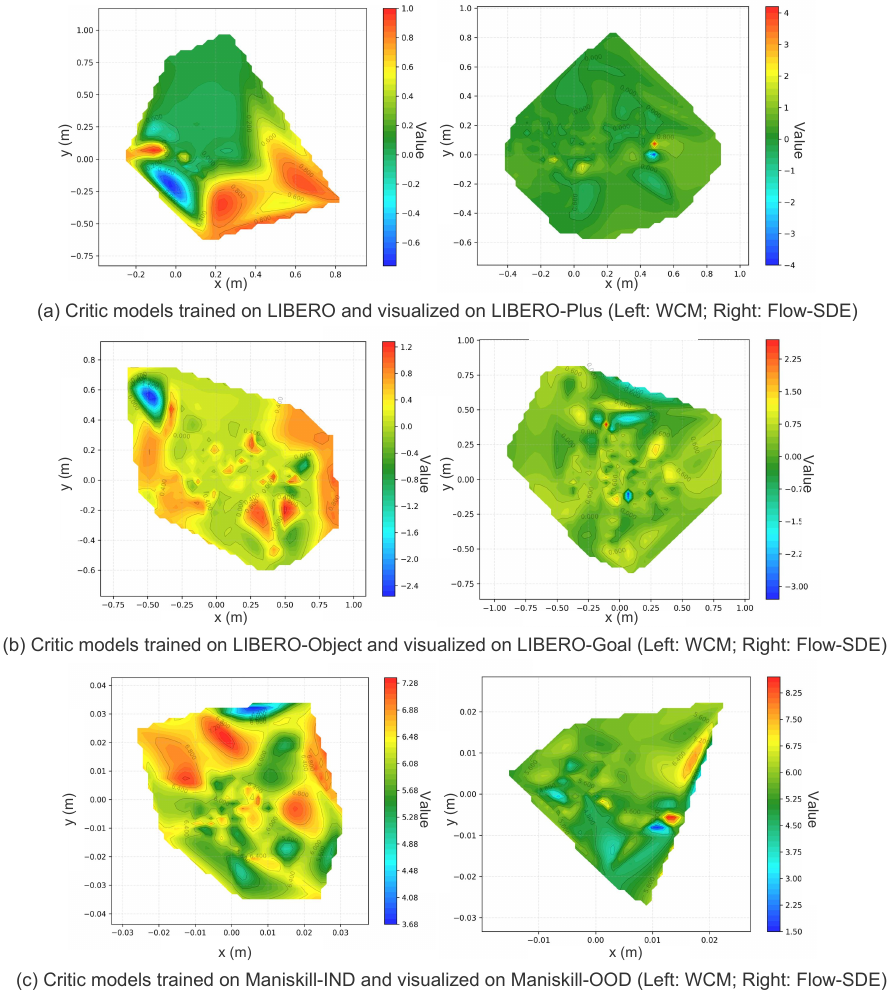}
    \caption{Heatmap of values predicted by the critic model at different coordinates within the robot’s reachable workspace. Warmer colors (red) indicate higher values.}
    \label{fig:heatmap}
    \vspace{-15pt}
\end{figure}

Specifically, we show three heatmaps (Figure~\ref{fig:heatmap}) that provide further substantiation for our findings. We randomly sample 1000 data points with different $x$ and $y$ coordinates within the robot's reachable space. For each sampled point, we save the corresponding observation and use a critic model to infer the value associated with that observation, and then generate a planar heatmap. We conduct three sets of experiments: (i) Training $\pi_{0.5}$ on LIBERO using Flow-SDE with WCM augmentation and the original Flow-SDE respectively for 200 steps, and then evaluating on the LIBERO-Plus benchmark; (ii) Training $\pi_{0.5}$ on the LIBERO Object Suite using Flow-SDE with WCM augmentation and the original Flow-SDE respectively for 200 steps, and then evaluating on the LIBERO Goal Suite; (iii) Training $\pi_{0.5}$ on Maniskill-IND using Flow-SDE with WCM augmentation and the original Flow-SDE respectively for 200 steps, and then evaluating on Maniskill-OOD.

We find that after scene generalization, the original method exhibits a decrease in value discriminability, while also being prone to generating excessively large or small values at a few points. This is because the critic model overfits to the original distribution and produces outliers when encountering out-of-distribution (OOD) samples. In contrast, the values produced by WCM retain a certain degree of discriminability and are less susceptible to local outliers, demonstrating stronger generalization ability.

\end{document}